\documentclass{article}

\usepackage{arxiv}

\usepackage[utf8]{inputenc} % allow utf-8 input
\usepackage[T1]{fontenc}    % use 8-bit T1 fonts
\usepackage{hyperref}       % hyperlinks
\usepackage{url}            % simple URL typesetting
\usepackage{booktabs}       % professional-quality tables
\usepackage{amsfonts}       % blackboard math symbols
\usepackage{nicefrac}       % compact symbols for 1/2, etc.
\usepackage{microtype}      % microtypography
\usepackage{lipsum}
\usepackage{graphicx}
\usepackage{natbib}
\usepackage{bm}
\usepackage{tikz}
\usetikzlibrary{arrows.meta, positioning, fit, decorations.pathreplacing, calc}
\usepackage{makecell}
\usepackage{xcolor}
\usepackage{multirow}
\usepackage{amsmath}

\usepackage[T1]{fontenc}
\usepackage[utf8]{inputenc}

\title{Ensuring Semantics in Weights of Implicit Neural Representations through the Implicit Function Theorem}

\author{%
  Tianming Qiu\textsuperscript{1,2}, ~Christos Sonis\textsuperscript{1}\thanks{Work done during an internship at Research Institute of the Free State of Bavaria for software-intensive
systems (fortiss).}, ~Hao Shen\textsuperscript{1,2} \\
\\
  \textsuperscript{1}Technical University of Munich\\ 
  \textsuperscript{2}Research Institute of the Free State of Bavaria
for software-intensive systems (fortiss)\\
  \texttt{\{tianming.qiu, christos.sonis, hao.shen\}@tum.de}}

\begin{document}
\maketitle
\begin{abstract}
Weight Space Learning (WSL), which frames neural network weights as a data modality, is an emerging field with potential for tasks like meta-learning or transfer learning.
Particularly, Implicit Neural Representations (INRs) provide a convenient testbed, where each set of weights determines the corresponding individual data sample as a mapping from coordinates to contextual values.
So far, a precise theoretical explanation for the mechanism of encoding semantics of data into network weights is still missing.
In this work, we deploy the Implicit Function Theorem (IFT) to establish a rigorous mapping between the data space and its latent weight representation space.
We analyze a framework that maps instance-specific embeddings to INR weights via a shared hypernetwork, achieving performance competitive with existing baselines on downstream classification tasks across 2D and 3D datasets.
These findings offer a theoretical lens for future investigations into network weights.
\end{abstract}

% keywords can be removed

\section{Introduction}
% background and motivation:
% millions of model - similiar functions - weight can capture similar semantics 
With rapid development of Artificial Intelligence (AI), we are witnessing an explosion
in both the number and scale of neural network models.
Despite their remarkable generali\-zation capabilities, many of these models
are often trained on overlapping benchmarks, or designed to perform on simi\-lar tasks.
Arguably, models trained on similar data could share similar underlying relationships in terms of their learned weights.
Recent works, such as the ``model atlas'', leverage model weights from the
platform \emph{Hugging Face} to visualize relationships among individual models and entire populations~\citep{horwitz2025we}.
Building on this perspective, network weights are increasingly viewed as a structured data modality,
i.e., once trained, they are expected to encode semantics of the training data.

% History and related work
The concept of neural network weights being capable of encoding data semantics can be traced back to Bayesian neural networks, where weights capture a posteriori
distribution that reflects the model's learned semantics of data~\citep{gal2016dropout}.
The currently emerging Weight Space Learning (WSL) research line~\citep{schurholt2024neural} further consolidates this trend by treating weights as a stand-alone modality.
Recent works on large-scale models such as HyperDreamBooth~\citep{ruiz2024hyperdreambooth} and HyperTuning~\citep{phang2023hypertuning}, show that modifying only a small subset of weights generated by hypernetworks~\citep{ha2016hypernetworks} can substantially alter models' behavior.
Furthermore, model merging~\citep{yang2024model} aims to integrate knowledge across multiple models by fusing their weights for better performance.
These approaches suggest that a better understanding of the relationship between data and its weight-space representation could benefit modern neural models in tasks such as meta-learning~\citep{finn2017model}, style transfer~\citep{gatys2016image, karras2019style}, and efficient fine-tuning~\citep{houlsby2019parameter}.

Among various models, Implicit Neural Representations (INRs) provide a particularly well-suited framework for investigating this requirement.
By parameterizing data samples as coordinate-based functions~\citep{park2019deepsdf}, INRs leverage neural networks' universal approximation ability~\citep{hornik1989multilayer, kidger2020universal} to successfully reconstruct a wide variety of data.
Crucially, each data sample is associated with its own neural network, whose weights serve as an encoding of the underlying semantics.
This makes INRs a natural testbed for analyzing how individual samples are embedded into weight spaces~\citep{navon2023equivariant,zhou2023permutation}.

For weight-space representations to be suitable and reliable for downstream tasks, an identifiable correspondence between an individual data sample and its associated neural parameters is demanded.
Most existing works on INR weights space learning address the issue of the symmetry structure of network weights, focusing on permutation-invariant constructions to enable learning over weights.
We argue that this difficulty arises not only from parameter symmetries, but also from the optimization process itself, where distinct local minima can correspond to functionally similar solutions.

More specifically, we investigate a jointly trainable ``auto-decoder''~\citep{park2019deepsdf} style hypernetwork-based INR architecture~\citep{sitzmann2019scene}.
We refer to this architecture as the \emph{HyperINR} framework, in which the latent embeddings and hypernetwork are learned jointly, with the hypernetwork mapping each latent embedding to INR parameters.
By using HyperINR as a minimal, yet non-trivial setting, the latent embedding has no explicit correspondence to the data and can be interpreted as a low-dimensional representation of network weights.
Examining the HyperINR loss through the lens of the \textit{Implicit Function Theorem}, we formulate the training process as an implicit relation defined by the reconstruction objective, and analyze when this relation admits a well-defined data-to-weight mapping.

In our experiments across 2D data and 3D shape classification tasks, we observe that the jointly optimized embeddings form distinct clusters in both latent and weight spaces.
Our outperforming classification accuracy further verifies numerically that the latent weight embeddings faithfully capture these semantics.
Compared to prior works, our approach employs \emph{minimal data pre-processing} 
and a \emph{parsimonious pipeline}, which highlights the potential effectiveness of this proof-of-concept paradigm.

In summary, our contributions are threefold:
(1) A precise mathematical description of existing HyperINR frameworks through the lens of the \emph{Implicit Function Theorem}.
% (2) A rigorous analysis of the conditions under which the IFT establishes a connection between data semantics and latent weight representations.
(2) A rigorous analysis showing that the full-rank condition
of the Jacobian of the corresponding implicit function ensures the faithful preservation 
of data semantics in the latent weight space.
(3) Experiments on 2D and 3D datasets demonstrate consistent improvements over existing INR-based weight classifiers.

\section{Related Work}
\label{sec:related_work}

\paragraph{INR weights classification.}
Using INR weights directly for downstream tasks, such as classification, can be challenging, since the
optimized INRs can vary significantly depending on their initializations, even for the same signal~\citep{huang2025few}.
Permutation-equivariant classifiers, such as DWSNet~\citep{navon2023equivariant, navon2024equivariant} and NFN~\citep{zhou2023permutation},
address these parameter symmetry issues by ensuring that models are invariant to weight reordering.
Similarly, Transformer-based architectures~\citep{zhou2023neural} can also produce permutation-equivariant representations of weights for downstream tasks.
More recently, MWT~\citep{gielisse2025end} introduces equivariant INR classifiers with end-to-end supervision, achieving strong performance, but in a setting where labels are integrated into the representation.
While these approaches make progress in handling permutation-induced symmetry, a more fundamental challenge remains unresolved.
Namely, optimizing neural networks even for the same data samples cannot be guaranteed to converge to the
same set of weights.
This reflects a fundamental non-uniqueness of over-parameterized INR weights that arises from the optimization process itself, beyond permutation symmetry.
In practice, this often leads symmetry-aware methods to depend on more intensive sampling of equivalent weight configurations, or increased model capacity to achieve invariance.

\paragraph{Weight space learning and its latent representation.}
Functa~\citep{dupont2022data} is a pioneering work that formalizes INR weights as data.
Namely, a shared base network captures common semantics, while per-sample modulation vectors encode variations.
These low-dimensional modu\-lation vectors are shown to be effective representations for downstream tasks.
Schürholt et al.~\cite{schurholt2022hyper} further learn hyper-representations via self-supervised training on the weights of INR model zoos, yielding embeddings that gene\-ralize across tasks.
Similarly, inr2vec~\citep{luigi2023deep} maps INR weights into latent embeddings for classification and generation, showing that meaningful semantic structure is preserved in weight space representations.
Another related work, ProbeGen~\citep{kahana2025deep}, optimizes latent embeddings to generate probes for classifying trained INRs.
These works indicate a growing interest and challenge in directly analyzing neural network weights and in discovering compact latent representations that capture their underlying semantics.

\paragraph{Auto-decoder and HyperINRs architecture.}
Early work on implicit neural representations (INRs), such as SIREN~\citep{sitzmann2020implicit}, already explores the use of hypernetworks~\citep{ha2016hypernetworks}, in which network weights are generated from latent embeddings conditioned on inputs.
Spurek et al.~\cite{pmlr-v119-spurek20a} use a hypernetwork to generate 3D point clouds.
DeepSDF~\citep{park2019deepsdf} introduces an auto-decoder architecture that learns instance-specific latent representations.
Similarly, SRNs~\citep{sitzmann2019scene} combine jointly trainable latent embeddings with a hypernetwork to reconstruct 3D scenes and objects.
More recently, D’OH~\citep{gordon2024d} proposed a decoder-only hypernetwork formulation, where instance embeddings are directly optimized from random initializations, without imposing structural assumptions on the latent space.
From an architectural perspective, a growing body of work adopts HyperINR formulations to improve generalization and reconstruction quality across instances and tasks~\citep{gu2023generalizable, gu2025foundation, chen2022transformers, kim2023generalizable}.

%

% \section{Methodology}
% \label{sec:methodology}
%
%In this section, we first introduce the HyperINR model in Section~\ref{sec:31}. Then in Section~\ref{sec:32}, 

% \subsection{A mathematical description of HyperINR frameworks}
\section{Preliminaries: a mathematical modeling of HyperINR frameworks}
\label{sec:31}
\begin{figure}[t!]
    \centering
    \input{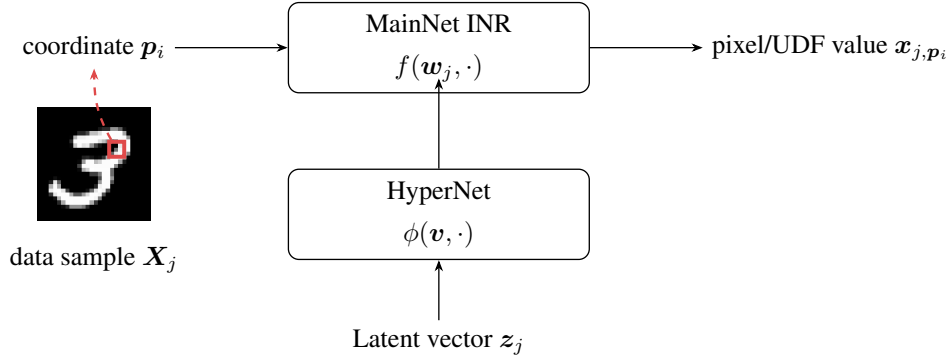}
    \caption{HyperINR: A hypernetwork generates weights $\bm{w}_j$ from learnable latent vector $\bm{z}_j$, which are then used by the main network $f(\bm{w}_j, \cdot)$ to map coordinates to pixel or Unsigned Distance Function (UDF) values.}
    \label{fig:HyperINR-architecture}
\end{figure}

Instead of learning low-dimensional representations from the weights of 
individually pretrained INRs for all data samples, a generic HyperINR 
framework deploys a hypernetwork to map a learnable low-dimensional latent space to the weight
space of INR main networks. The HyperINR pipeline is depicted in 
Fig.~\ref{fig:HyperINR-architecture}.
In what follows, we model this generic HyperINR framework in a mathematical 
concise manner to facilitate our analysis in the following section.

Let $\mathcal{P} \subset \mathbb{R}^{p}$ be an open set of
$p$-dimensional coordinates and $\mathcal{W} \subset \mathbb{R}^{d}$ be the set of weights of INRs.
An INR main network takes sampled coordinates $\bm{p} \in \mathcal{P}$ as input and
generates the corresponding pixel value or Unsigned Distance Function (UDF) value~\citep{mullen2010signing}
in $\mathbb{R}^c$.
Specifically, for a set of sampled coordinates $\{\bm{p}_i\}_{i=1}^n \subset \mathcal{P}$, any data example
$\bm{X} := \{\bm{x}_{i}\}_{i=1}^{n}\in \mathcal{X} \subset \mathbb{R}^{n \times c}$ can be represented as
a sampled mapping from coordinate $\bm{p}_{i}$ to data value $\bm{x}_{i}$.
Here, we define an INR main network as the following mapping
\begin{equation}
    f \colon \mathcal{W} \times \mathcal{P} \to \mathbb{R}^{c},
    \qquad
    (\bm{w}, \bm{p}) \mapsto f(\bm{w}, \bm{p}).
    % f \colon \mathbb{R}^{d} \times \mathbb{R}^{p} \to \mathbb{R}^{c},
    % \qquad
    % (\bm{w}, \bm{p}) \mapsto f(\bm{w}, \bm{p}),
\end{equation}
In this work, we assume that the set of weights of INRs admits a low-dimensional structure.
Specifically, the weights $\bm{w}_i$ for each data sample are generated by a hypernetwork, which maps
a learnable latent embedding $\bm{z}_j \in \mathcal{Z} \subset \mathbb{R}^{l}$ to the weights
$\bm{w}_i \in \mathcal{W}$ of the main network, i.e.,
\begin{equation}
    \phi \colon \mathcal{V} \times \mathcal{Z} \to \mathcal{W},
    \qquad
    (\bm{v}, \bm{z}_j) \mapsto \phi(\bm{v}, \bm{z}_j),
\end{equation}
where $\mathcal{V} \subset \mathbb{R}^{k}$ denotes the space of hypernetwork weights, which are shared across
all data examples.
For a given set of data samples $\{\bm{X}_{j}\}_{j=1}^{t}$, a common training strategy is to find
an optimal set of weights of the hypernetwork $\bm{v}^{*} \in \mathcal{V}$ and a set of latent embeddings
$\{ \bm{z}_{j}^{*} \}_{j=1}^{t} \subset \mathcal{Z}$, such that the corresponding INRs exactly represent the data samples, i.e.,
for all $i = 1, \ldots, n$ and $j = 1, \ldots t$, the following equation holds true
\begin{equation}
    f( \phi( \bm{v}^{*}, \bm{z}_{j}^{*} ), \bm{p}_{i} ) = \bm{x}_{j,\bm{p}_{i}},
\end{equation}
with $\bm{X}_{j} := \{\bm{x}_{j,i}\}_{i=1}^{n}\in \mathcal{X}$.
With such a construction, it is clear that the latent embeddings $\bm{z}_{j}^{*}$ can be considered as a
point-wise representation of the original data samples $\bm{X}_{j}$.
The core question investigated in this work is how to ensure that the semantics of data are captured
in the weights of corresponding INRs.

% \subsection{Ensuring semantics in weights with the Implicit Function Theorem}
\section{Analysis of HyperINR with the IFT}
%Global mapping between data and weight representations}}
%
% Weight space can be viewed as a low-dimensional manifold, and we use a latent embedding $\bm{z}$ to capture its structure. The high dimensionality of weights contains redundancies, not only due to permutation symmetries but also from the non-uniqueness of local minima in the optimization landscape. To formalize the relationship between data and weights, we leverage the Implicit Function Theorem (IFT).

% Let us define an INR mainnet as
% \begin{equation}
%     f \colon \mathbb{R}^{d} \times \mathbb{R}^{p} \to \mathbb{R}^{c},
%     \qquad
%     (\bm{w}, \bm{p}) \mapsto f(\bm{w}, \bm{p}),
% \end{equation}
% where $\bm{w} \in \mathbb{R}^d$ are the weights, $\bm{p} \in \mathbb{R}^p$ are input coordinates, and the output is in $\mathbb{R}^c$.
% %
% Each data example $\bm{X}_i \in \mathbb{R}^{n \times c}$ is associated with a set of coordinates $\{\bm{p}_j\}_{j=1}^n$.
% The hypernetwork
% \begin{equation}
%     \phi: \mathbb{R}^k \times \mathbb{R}^{\ell} \to \mathbb{R}^d,
%     \qquad
%     (\bm{v}, \bm{z}_i) \mapsto \phi(\bm{v}, \bm{z}_i),
% \end{equation}
% maps a latent embedding $\bm{z}_i \in \mathbb{R}^{\ell}$ to the weights $\bm{w}_i \in \mathbb{R}^{d}$ of the main network, while the global parameters $\bm{v} \in \mathbb{R}^{k}$ are the hypernetwork weights shared across all data examples.
%
% text for section 3.2
%
In this section, we first investigate the residual loss
for training the HyperINR, and then apply the IFT to analyze
the global minimum of the loss function to ensure the capture of the semantics of data
into weights of INRs.

In our analysis, we assume that the data set $\mathcal{X} \subset \mathbb{R}^{n \times c}$
is a compact local-dimensional manifold, and the space of latent embeddings $\mathcal{Z} \subset
    \mathbb{R}^{l}$ is an open set.
The classic residual loss for assessing the reconstruction of the
INR against a given data sample can be defined as
\begin{equation}
    \begin{split}
        \ell \colon \mathcal{V} \times \mathcal{Z} \times
        \mathcal{X}                     & \to \mathbb{R},
        \\
        \ell(\bm{v}, \bm{z}, \bm{X}) := & \, \tfrac{1}{2} \Big\| \big\{ f(\phi(\bm{v}, \bm{z}), \bm{p}_{i}) \big\}_{i=1}^{n} - \bm{X} \Big\|_{F}^{2} \\
        =                               & \, \tfrac{1}{2} \sum\limits_{i=1}^{n} \Big\| f(\phi(\bm{v}, \bm{z}), \bm{p}_{i})
        - \bm{x}_{\bm{p}_{i}} \Big\|_{2}^{2},
    \end{split}
\end{equation}
where $\bm{x}_{\bm{p}_i}$ denotes the value of sample $\bm{X}$ at coordinate $\bm{p}_i$, and
$\|\cdot\|_{F}$ denotes the Frobenius norm of matrices.
Let us assume that the size of the INRs is sufficiently large to ensure exact reconstruction in
the whole data manifold $\mathcal{X}$, i.e., there is a hypernetwork weight $\bm{v}^* \in \mathcal{V}$,
which ensures exact reconstruction of any sample $\bm{X} \in \mathcal{X}$.
Consequently, for any pair $(\bm{z},\bm{X}) \in \mathcal{Z} \times \mathcal{X}$,
the gradient of $\ell$ vanishes at $(\bm{v}^{*}, \bm{z}, \bm{X})$.

With a fixed $\bm{v}^*$ for the hypernetwork, we define the following function
\begin{equation}
    \ell_{\bm{v}^*} \colon \mathcal{Z} \times \mathcal{X} \to \mathbb{R},
    \qquad
    (\bm{z}, \bm{X}) \mapsto \ell(\bm{v}^*, \bm{z}, \bm{X}).
    %
    % \ell_{\bm{v}^*}(\bm{z}_i, \bm{X}_i) := \ell(\bm{v}^*, \bm{z}_i, \bm{X}_i).
\end{equation}
Let us denote $\bm{w} := \phi(\bm{v}^{*},\bm{z})$, and the pixel-wise residual by
\begin{equation}
    \bm{\epsilon}_{i} := f(\phi(\bm{v}^*, \bm{z}), \bm{p}_{i}) - \bm{x}_{\bm{p}_{i}}.
\end{equation}
We compute the differential map of $\ell_{\bm{v}^*}$ with respect to the
hyperparameter $\bm{z}$ as
\begin{equation}
    \operatorname{D}_{1} \ell_{\bm{v}^*}(\bm{z}, \bm{X})
    = \sum_{i=1}^n \bm{\epsilon}_{i}^\top \operatorname{D}_{1}f(\bm{w},\bm{p}_{i})
    \operatorname{D}_{2} \phi(\bm{v}^{*}, \bm{z}),
\end{equation}

where $\operatorname{D}_{1}f(\bm{w},\bm{p}_{i})$ is the differential map of $f$ with respect to
the first variable, and $\operatorname{D}_{2} \phi(\bm{v}^{*}, \bm{z})$ is the differential map
of $\phi$ with respect to the second variable.
For any pair $(\bm{z},\bm{X}) \in \mathcal{Z} \times \mathcal{X}$, we have the vanishing
gradient as
\begin{equation}
    \operatorname{D}_{1} \ell_{\bm{v}^*}(\bm{z}, \bm{X}) = 0.
\end{equation}
Clearly, the above equation establishes a potential mapping between
$\bm{z}$ and $\bm{X}$ in terms of implicit functions.

Let us define the following function
\begin{equation}
    \label{eq:if}
    \xi_{\bm{v}^{*}} \colon \mathcal{Z} \times
    \mathcal{X} \to
    \mathbb{R}^{l}, \quad
    (\bm{z}, \bm{X}) \mapsto ( \operatorname{D}_{1} \ell_{\bm{v}^*}(\bm{z}, \bm{X}) )^{\top},
\end{equation}
which is essentially the gradient of $\ell_{\bm{v}^{*}}$ with respect to $\bm{z}$, by endowing the
canonical Euclidean metric.
We then investigate the properties of the Jacobian of $\xi_{\bm{v}^{*}}$ with
respect to the first variable at $\bm{z}$, which is nothing but the
second differential of $\ell_{\bm{v}^{*}}$ with respect to $\bm{z}$.
Straightforwardly, we have
%
% Let us compute the derivative of the differential map
% $\operatorname{D}_{1} \ell_{\bm{v}^*}(\bm{z}_i^*, \bm{X}_i)$
% with respect to $\bm{z}^{*}$ as
%
% \begin{equation}
% \label{eq:hessian}
% \begin{split}
%     \operatorname{D}_{1} \xi_{\bm{v}^{*}}(\bm{z},\bm{X}) = 
%     \sum\limits_{i=1}^{n}
%     \left( \operatorname{D}_{1}f(\bm{w},\bm{p}_{i})
%         \operatorname{D}_{2} \phi(\bm{v}^{*}, \bm{z}) \right)^{\top}
%     \operatorname{D}_{1}f(\bm{w},\bm{p}_{i})
%         \operatorname{D}_{2} \phi(\bm{v}^{*}, \bm{z}).
%         \end{split}
% \end{equation}
%
\begin{equation}
    \label{eq:hessian}
    \begin{aligned}
        \operatorname{D}_{1} \xi_{\bm{v}^{*}}(\bm{z},\bm{X})
        = \sum_{i=1}^{n}
         & \Bigl(
        \operatorname{D}_{1}f(\bm{w},\bm{p}_{i})
        \operatorname{D}_{2} \phi(\bm{v}^{*}, \bm{z})
        \Bigr)^{\top} \cdot
        \operatorname{D}_{1}f(\bm{w},\bm{p}_{i})
        \operatorname{D}_{2} \phi(\bm{v}^{*}, \bm{z}) .
    \end{aligned}
\end{equation}
Note, that the above expression is essentially the Hessian matrix of the
residual loss $\ell$ with respect to the exact
latent representation $\bm{z}$
of each sample $\bm{X}$ at a globally optimal hypernetwork weight $\bm{v}^{*}$.
It is known that if $\operatorname{D}_{1} \xi_{\bm{v}^{*}}(\bm{z},\bm{X})$ is of full rank
on $\mathcal{Z} \times \mathcal{X}$, then the value of zero is a regular value
of the map $\xi_{\bm{v}^{*}}$.
Furthermore, the global implicit function theorem~\citep{krantz2002implicit} implies that there is a
unique smooth map $g \colon \mathcal{X} \to \mathcal{Z}$, which satisfies
for all $\bm{X} \in \mathcal{X}$
\begin{equation}
    \label{eq:ift}
    \xi_{\bm{v}^{*}} (g(\bm{X}), \bm{X}) = 0.
\end{equation}
\vspace{-3mm}

By knowing $\operatorname{D}_{1}f(\bm{w},\bm{p}_{i}) \in \mathbb{R}^{c \times d}$,
$\operatorname{D}_{2} \phi(\bm{v}^{*}, \bm{z}) \in \mathbb{R}^{d \times l}$, and
$c < l$, we see that the Jacobian of $\xi_{\bm{v}^{*}}$ is a sum of
$n$ positive semi-definite symmetric matrices with their ranks not larger than $c$.
Therefore, a necessary condition to ensure the Jacobian of $\xi_{\bm{v}^{*}}$
to have full rank of $l$ is $nc \geq l$.
This simple condition can be used as a guidance to determine the minimal number of coordinates
that is needed to ensure the semantics of data are captured in the weights of the
corresponding INRs.
Note, that this is typically satisfied for both 3D shape and image data, where the dimension
of latent vector $l$ is always much smaller than $nc$.
Moreover, it is also worth noticing that the rank of the Jacobian of
$\xi_{\bm{v}^{*}}$ depends further on various factors, such as the geometry of the data manifold
$\mathcal{X}$, the architectures of both the main network and the hypernetwork, and
the choice of activation functions.
Thus, the investigation of sufficient conditions for the Jacobian being of full rank is
considered to be a challenging research question in the future.

As shown in our analysis above, by leveraging the gradient of the residual loss $\ell$
as implicit constraints between the data manifold $\mathcal{X}$ and
the latent space of weights of INRs $\mathcal{Z}$, global semantics of the data manifold can
be captured in the latent space of INR weights, under the conditions of exact reconstruction of INRs.
Practically, for a given finite number of samples $\{ \bm{X}_{j} \}_{j=1}^{t} \subset \mathcal{X}$, we define
the total residual loss function of all data samples as
\begin{equation}
    \mathcal{L}(\bm{v}, \{\bm{z}_{j}\}_{j=1}^t) := \sum_{j=1}^t \ell(\bm{v}, \bm{z}_{j}, \bm{X}_{j}).
\end{equation}
By minimizing the total residual loss, we assume a hypernetwork weight
$\bm{v}^{*}$ that reconstruct exactly all the samples.
If the Jacobian as in Eq.~\eqref{eq:hessian} at all samples has full
rank, the classic IFT implies that there exist open neighborhoods $\mathcal{N}_{\bm{X}_{j}}$ containing $\bm{X}_{j}$ and
$\mathcal{N}_{\bm{z}_{j}^{*}}$ containing $\bm{z}_{j}^{*}$, such that
there is a unique continuously differentiable function
$g \colon \mathcal{N}_{\bm{X}_{j}} \to \mathcal{N}_{\bm{z}_{j}^{*}}$, which satisfies
\begin{equation}
    %\label{eq:ift}
    \begin{split}
         & g(\bm{X}_{j})=\bm{z}_{j}^{*}, \quad \text{and}              \\
         & \operatorname{D}_{1} \xi_{\bm{v}^*}(g(\bm{X}), \bm{X}) = 0,
        \text{ for all } \bm{X} \in \mathcal{N}_{\bm{X}_{j}}.
    \end{split}
\end{equation}
Clearly, only local semantics around the samples is guaranteed to be
captured in the neighborhood
around their corresponding latent embeddings.
Nevertheless, we hypothesize that a sufficiently large number of samples,
which uphold the conditions for the local IFT,
%the constraint on training the HyperINR model, which enforces the bijective mapping between a sample $\bm{X}_{j}$ and its latent representation $\bm{z}_{j}^{*}$ for all $j = 1, \ldots, t$,
enhance the chance of global semantics of data being preserved in the latent space of INR weights.
Our experimental results in Section~\ref{sec:53} explore further this hypothesis.

\section{Experiments}
\label{sec:experiments}

\subsection{Implementation and Joint Optimization Protocol}
\begin{figure}[t!]
    \centering
    \input{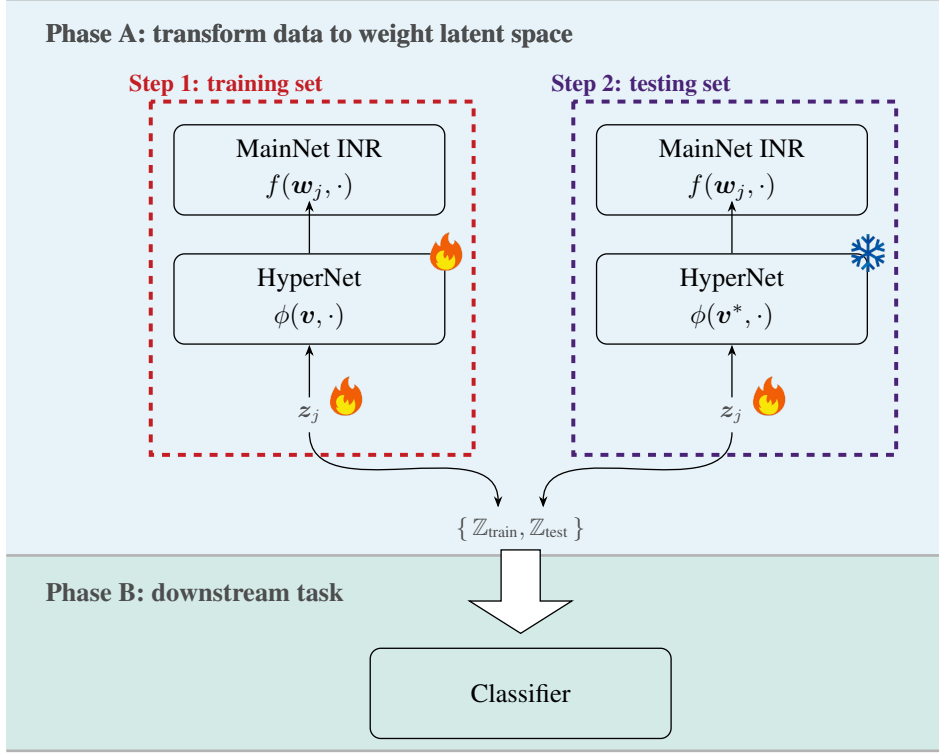}
    \caption{Phase A transforms data samples to latent weight representations using HyperINR, while Phase B utilizes these latent representations for downstream classification tasks. The fire symbol indicates parameters that are updated, while the snowflake symbol indicates parameters that are frozen.}
    \label{fig:pipeline}
\end{figure}
The common paradigm for INR weight classifiers and ``weight as a modality'' works typically involves two phases: first, training INRs to reconstruct data samples to transform semantics in the raw input data to INR weights or latent weight space; second, using the optimized weights and latent weight representations for downstream tasks.
We also follow this two-phase training pipeline shown in Fig.~\ref{fig:pipeline}.
The Phase A transformation process, which prepares the weight data, is performed in two steps: one for the training set and one for the testing set.
In the first step, we prepare the training set by jointly optimizing the hypernetwork $\phi(\bm{v}, \cdot)$ and the latent embeddings $\{\bm{z}_i\}$ to minimize the reconstruction loss for all training samples.
The latent embeddings are initialized randomly around the origin and updated during the optimization.
The second step involves fixing weights $\bm{v}^*$ of the hypernetwork and inferring new embeddings for unseen test data by minimizing the reconstruction loss.
Prior approaches require a separate stage for fitting the INRs and then transforming it to a low-dimensional space~\citep{luigi2023deep, zhou2023permutation}.
Our joint optimization of INR reconstruction and embedding is inherently more parallel-efficient.

To evaluate the effectiveness of our HyperINR framework and the learned latent embedding of weights, we conduct classifications on both 2D images and 3D shapes, following prior works~\citep{navon2023equivariant,zhou2023permutation,zhou2023neural,luigi2023deep} in Phase B.
We evaluate on five datasets: MNIST~\citep{yann2010mnist} and FashionMNIST~\citep{xiao2017fashion} for 2D images, and ModelNet40~\citep{wu20153d}, ShapeNet10~\citep{chang2015shapenet}, and ScanNet10~\citep{qin2019pointdan} for 3D shapes.
More dataset details are provided in Appendix~\ref{sec:appendix-dataset}.

\subsection{Clustering and Classification}
\begin{figure}[ht]
    \centering
    \includegraphics[width=1\linewidth]{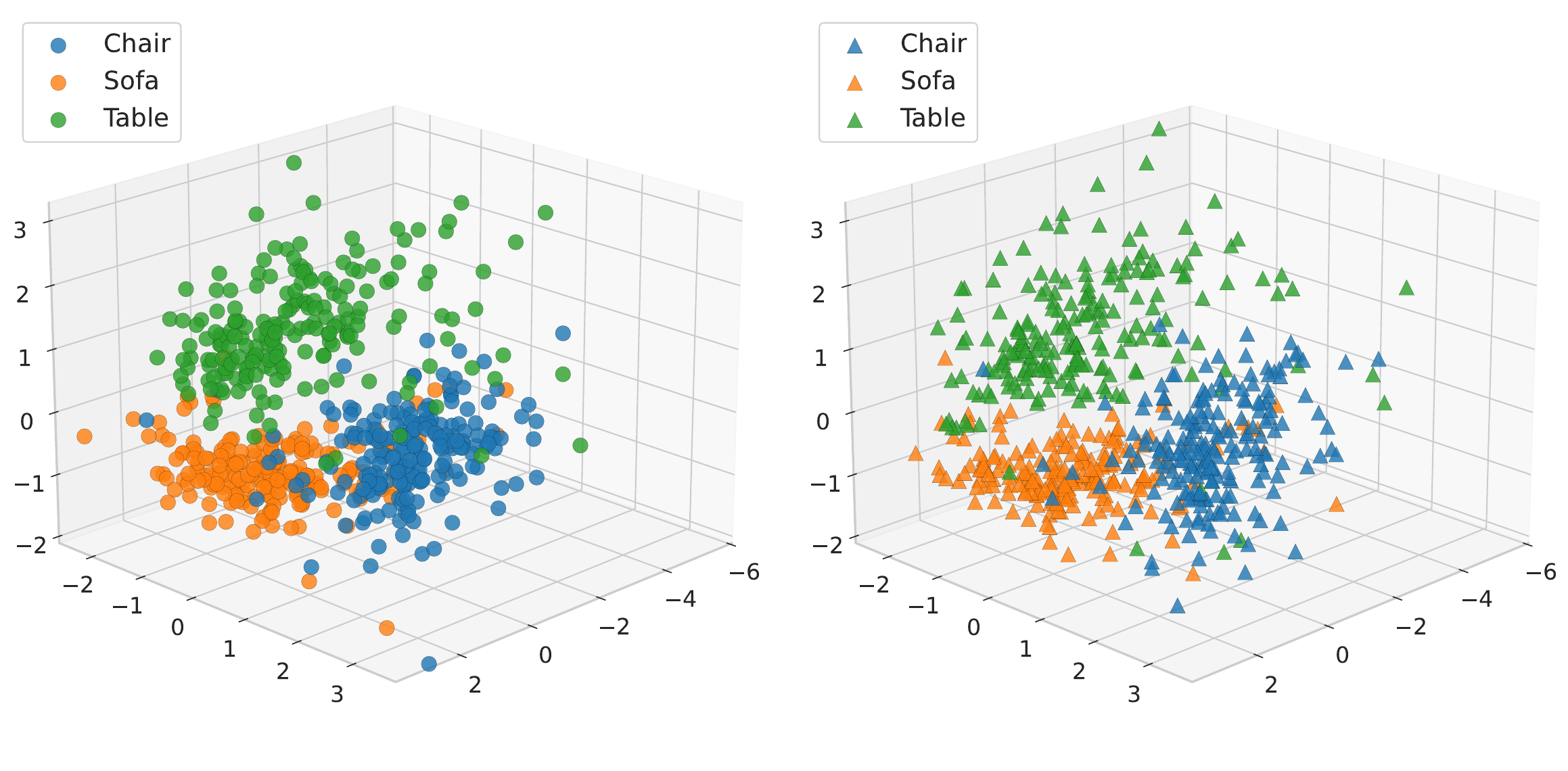}
    \caption{Distinct clustering of latent embeddings $\bm{z}$ from ShapeNet10 using 3D PCA. \textbf{Left:} Training samples form well-separated clusters. \textbf{Right:} Test samples fall into the neighborhoods of their respective categories under a fixed hypernetwork.}
    \label{fig:latent_z}
\end{figure}
\begin{figure}[ht]
    \centering
    \includegraphics[width=0.8\linewidth]{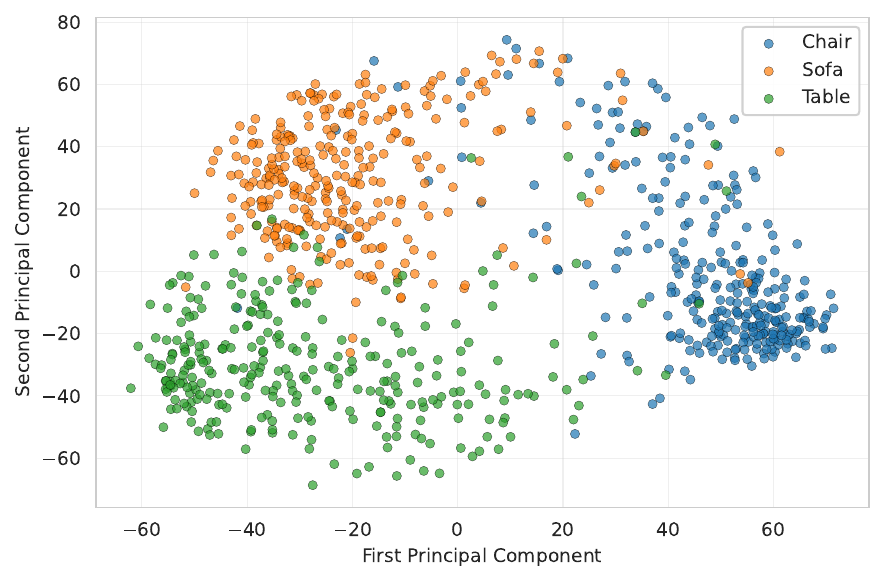}
    \caption{Distinct clustering of hypernetwork-generated weights for ShapeNet10 using 2D PCA.}
    \label{fig:latent_w}
\end{figure}

\paragraph{Clustering in weight- and latent space.} We first examine whether the learned latent embeddings exhibit clustering aligned with semantic classes.
Using ShapeNet10 as an example, we train on all ten classes but visualize embeddings from three categories (chair, sofa, table) for clarity.
For visualization, we apply Principal Component Analysis (PCA)~\citep{jolliffe2002pca}, a linear dimensionality reduction method, to avoid distortions introduced by nonlinear projections.
As shown in Figure~\ref{fig:latent_z}, the training embeddings form well-separated clusters, and test samples fall into the neighborhoods of their respective categories with the hypernetwork fixed.
Such distinct clustering is already visible under our minimal setup.
Using only the mean squared error (MSE) loss, a low-dimensional latent weight space assumption, and the guarantees provided by the IFT, the latent embeddings appear to preserve semantic information.
Furthermore, PCA projections of the hypernetwork-generated weights exhibit similarly distinct clusters, reinforcing that semantics are preserved in both latent and weight spaces, as shown in Fig.~\ref{fig:latent_w}.
Achieving similarly clean and well-separated clusters directly in weight space has been challenging in prior work, suggesting that our lightweight setup captures semantic structure more explicitly.

\begin{table}[t]
    \caption{
        Classification top-1 accuracy (\%) on 2D datasets.
        Results for NFN* and Inr2Array* are reproduced by us under a matched comparison protocol (dataset enlargement using $10\times$ data augmentation is removed).
        Results for inr2vec-arch are taken from the re-implementation of \cite{navon2023equivariant}, while other baselines are reported from the original papers.}
    %\textcolor{red}{The “No-Extras” Inr2Array reproduced by us is trained without warmup steps, weight decay, or data augmentations. The “No-Extras” NFN resuts are taken from~\citep{kalogeropoulos2024scale}, while the inr2vec results are obtained without pre-training and are taken from~\citep{navon2023equivariant}.}
    \label{tab:classification-results-2d}
    \begin{center}

        \begin{small}
            \begin{sc}

                \begin{tabular}{lcc}
                    \toprule
                    \multicolumn{1}{c}{Method} &
                    \multicolumn{1}{c}{MNIST}  &
                    \multicolumn{1}{c}{FashionMNIST} \\
                    \midrule

                    inr2vec-arch               &
                    23.69 $\pm$ 0.10           &
                    22.33 $\pm$ 0.41                 \\

                    DWSNet                     &
                    85.71 $\pm$ 0.57           &
                    67.06 $\pm$ 0.29                 \\

                    NFN                        &
                    92.9 $\pm$ 0.218           &
                    75.6 $\pm$ 1.07                  \\

                    % NFN (reproduced by~\cite{kalogeropoulos2024scale}) &
                    % \textcolor{red}{79.11 $\pm$ 0.84} &
                    % \textcolor{red}{68.94 $\pm$ 0.64} \\

                    Inr2Array                  &
                    \textbf{98.5 $\pm$ 0.00}   &
                    79.3 $\pm$ 0.00                  \\
                    \midrule
                    NFN*                       &
                    61.27 $\pm$ 0.00           &
                    57.29 $\pm$ 0.00                 \\

                    Inr2Array*                 &
                    76.93 $\pm$ 0.00           &
                    67.77 $\pm$ 0.00                 \\
                    \midrule

                    % Inr2Array (reproduced by us) &
                    % \textcolor{red}{98.28} &
                    % -- \\

                    % Inr2Array w/o reg. or aug. (reproduced by us) &
                    % \textcolor{red}{74.30} &
                    % -- \\

                    \textbf{Ours-z}            &
                    \textbf{97.89 $\pm$ 0.29}  &
                    \textbf{86.82 $\pm$ 0.98}        \\
                    \textbf{Ours-w}            &
                    \textbf{97.65 $\pm$ 0.23}  &
                    \textbf{86.50 $\pm$ 1.37}        \\
                    \bottomrule
                \end{tabular}
            \end{sc}
        \end{small}

    \end{center}
\end{table}

\begin{table}[t]
    \caption{Classification top-1 accuracy (\%) on 3D datasets. Results for NFN~\citep{zhou2023permutation} and inr2vec~\citep{luigi2023deep} are reported as in the original papers. }

    \label{tab:classification-results-3d}
    \begin{center}
        \begin{small}
            \begin{sc}

                \begin{tabular}{lccc}
                    \toprule
                    \multicolumn{1}{c}{Method} & \multicolumn{1}{c}{ModelNet40} & \multicolumn{1}{c}{ShapeNet10} & \multicolumn{1}{c}{ScanNet10} \\
                    \midrule
                    NFN                        & -                              & 88.7 $\pm$ 0.461               & 65.9 $\pm$ 1.10               \\
                    inr2vec                    & 87.0                           & 93.3                           & \textbf{72.1}                 \\
                    \midrule
                    \textbf{Ours-z}            & \textbf{87.61 $\pm$ 0.21}      & \textbf{94.27 $\pm$ 0.28}      & \textbf{69.77 $\pm$ 0.73}     \\
                    \textbf{Ours-w}            & \textbf{87.20 $\pm$ 0.40}      & \textbf{94.31 $\pm$ 0.07}      & \textbf{69.64 $\pm$ 0.30}     \\
                    \bottomrule
                \end{tabular}
            \end{sc}
        \end{small}
    \end{center}
\end{table}

\paragraph{Classification in weight- and latent space.} We evaluate the learned representations using a simple MLP classifier on five datasets, with results summarized in Tables~\ref{tab:classification-results-2d} and~\ref{tab:classification-results-3d}.
All results from our approach are reported as ``mean $\pm$ standard error'' over five different random seeds; details on seed selection and experimental variability are provided in Appendix~\ref{app:ablation}.

Our method achieves state-of-the-art performance on FashionMNIST, ModelNet40, and ShapeNet10, while remaining competitive on MNIST and ScanNet10.
Table~\ref{tab:classification-results-2d} includes both results reported in prior work and results reproduced by us.
In particular, results for NFN* and Inr2Array* are reproduced under a simplified protocol by removing their dataset enlargement strategy, which increases the effective dataset size by using 10 different INRs per data point.
Results for inr2vec-arch are taken from the re-implementation of~\cite{navon2023equivariant} using the same architecture without pretraining, while results for DWSNet~\citep{navon2023equivariant}, NFN~\cite{zhou2023permutation}, and Inr2Array~\citep{zhou2023neural} are reported as in the original papers.
Detailed comparisons with all baselines are provided in Appendix~\ref{app:unifying_protocol}.

We report classification results using both the latent embeddings $\bm{z}$ and the generated INR weights $\bm{w}$ as inputs to the classifier.
Apart from the input dimensionality, the classifier architecture and capacity are kept identical.
The results indicate that $\bm{z}$ provides a compact and effective representation of $\bm{w}$, suggesting that the high-dimensional weight space is constrained to a low-dimensional structured manifold.
Based on this observation, all subsequent comparisons are conducted using the latent representations $\bm{z}$.

\subsection{Empirical Validation of IFT conditions}
\label{sec:53}
\begin{table}[t]
    \caption{Reconstruction and critical point gradient analysis. Average reconstruction error and average gradient norm with respect to the latent embedding $\bm{z}$ for both the embedding and hypernetwork methods.}
    \label{tab:reconstruction-error}
    \begin{center}
        \begin{small}
            \begin{sc}
                \begin{tabular}{lcccc}
                    \toprule
                                                & \multicolumn{2}{c}{ Reconstruction} & \multicolumn{2}{c}{ Latent vector}                               \\
                    \multicolumn{1}{c}{Dataset} & \multicolumn{2}{c}{error}           & \multicolumn{2}{c}{gradient norm}                                \\
                    \cline{2-3} \cline{4-5}
                    \\[-0.5em]
                                                & Step 1                              & Step 2                             & Step 1       & Step 2       \\
                    \midrule
                    MNIST                       & $2.27e^{-2}$                        & $3.08e^{-2}$                       & $7.14e^{-3}$ & $9.59e^{-4}$ \\
                    F-MNIST                     & $4.55e^{-2}$                        & $5.78e^{-2}$                       & $2.04e^{-3}$ & $9.67e^{-4}$ \\
                    ModelNet40                  & $7.46e^{-4}$                        & $1.88e^{-3}$                       & $2.30e^{-4}$ & $5.18e^{-5}$ \\
                    ShapeNet10                  & $7.27e^{-4}$                        & $1.21e^{-3}$                       & $1.15e^{-4}$ & $2.00e^{-5}$ \\
                    ScanNet10                   & $9.77e^{-4}$                        & $4.86e^{-3}$                       & $2.66e^{-4}$ & $8.69e^{-5}$ \\
                    \bottomrule
                \end{tabular}
            \end{sc}
        \end{small}
    \end{center}
\end{table}

\begin{table*}[t]
    \caption{Hessian conditioning statistics across datasets.
        Percentage of Hessian matrices with condition number above a given threshold or smallest singular value below a threshold, evaluated on training and test sets.}
    \label{tab:hessian-analysis}
    \begin{center}
        \begin{small}
            \begin{sc}
                \begin{tabular}{ccccccccccc}
                    \toprule
                    \multicolumn{1}{c}{Criterion}
                     & \multicolumn{2}{c}{MNIST}
                     & \multicolumn{2}{c}{FashionMNIST}
                     & \multicolumn{2}{c}{ModelNet40}
                     & \multicolumn{2}{c}{ShapeNet10}
                     & \multicolumn{2}{c}{ScanNet10}            \\
                    \cmidrule(lr){2-11}
                     & train                            & test
                     & train                            & test
                     & train                            & test
                     & train                            & test
                     & train                            & test  \\
                    \midrule
                    $k > 10^{3} (\%)$
                     & 16.42                            & 20.18
                     & 0.43                             & 0.85
                     & 16.91                            & 6.60
                     & 40.94                            & 58.20
                     & 0.59                             & 0.40  \\
                    $k > 10^{4} (\%)$
                     & 1.56                             & 2.17
                     & 0.04                             & 0.06
                     & 1.78                             & 0.12
                     & 2.69                             & 1.20
                     & 0.02                             & 0.00  \\
                    $\sigma_{min} < 10^{-5} (\%)$
                     & 0.00                             & 0.00
                     & 0.00                             & 0.00
                     & 1.71                             & 0.12
                     & 2.65                             & 1.06
                     & 0.02                             & 0.00  \\
                    $\sigma_{min} < 10^{-6} (\%)$
                     & 0.00                             & 0.00
                     & 0.00                             & 0.00
                     & 0.19                             & 0.04
                     & 0.27                             & 0.25
                     & 0.00                             & 0.00  \\
                    \bottomrule
                \end{tabular}
            \end{sc}
        \end{small}
    \end{center}
\end{table*}

It is essential to verify whether the conditions required by the IFT hold in practice on real data.
We therefore provide empirical evidence supporting the applicability of the IFT in our setting.
We first examine reconstruction errors and gradient norms.
According to Eq.~\eqref{eq:if} and Eq.~\eqref{eq:ift}, the implicit function is defined at the critical points of the loss with respect to the latent embeddings $\bm{z}$; hence, we verify that the average gradient norm of $\bm{z}$ is very close to zero.
Moreover, as indicated in Eq.~\eqref{eq:hessian}, the Hessian matrix is full rank only when all data samples are represented exactly.
As shown in Table~\ref{tab:reconstruction-error}, both the training and testing sets achieve sufficiently low reconstruction errors, which can be regarded as negligible.

While reconstruction errors on MNIST and FashionMNIST are slightly higher than those on other datasets, they remain lower than or comparable to those reported in prior work.
For instance, Inr2Array reports MSE reconstruction losses of $0.0270 \pm 0.0030$ on MNIST and $0.0700 \pm 0.0060$ on FashionMNIST, whereas our HyperINR achieves 0.0227 and 0.0455, respectively, demonstrating competitive reconstruction quality under a minimal training setup.

We further examine the Hessian matrix, which corresponds to the Jacobian of the implicit function.
Additional statistics on the Hessian conditioning distributions are provided in Appendix~\ref{app:hess}.
Here, we summarize the results by numerically analyzing the singular values and condition numbers of the Hessian.

As shown in Table~\ref{tab:hessian-analysis}, across all datasets, the Hessian is non-degenerate for the vast majority of samples, with only a small fraction exhibiting large condition numbers or near-zero singular values.
In particular, the proportion of Hessians with condition number exceeding $10^4$ is below $3\%$ across all datasets and splits, while occurrences of extremely small singular values ($\sigma_{\min} < 10^{-6}$) are rare or absent.
This indicates that the Hessian is not only generically full rank, but also reasonably well-conditioned in practice.

Overall, these results support the applicability of the IFT in practice: the Hessian is full rank for almost all samples, and pathological cases with severe ill-conditioning are statistically negligible.
This suggests that deviations from the IFT assumptions are unlikely to drive the observed behavior of our method.

Notably, the Hessian matrix here is the Jacobian matrix of the implicit function as given in Eq.~\eqref{eq:if}, namely the Hessian matrix of the reconstruction loss, and this behavior persists even though no explicit constraints or regularization are imposed on the loss Hessian matrix during training.
This suggests that the HyperINR framework implicitly induces well-conditioned implicit mappings.
We leave the explicit control or regularization of the Hessian, when required by specific tasks, as a direction for future work.

\subsection{Smooth Interpolations in Latent Space}
\begin{figure*}[ht]
    \centering
    \input{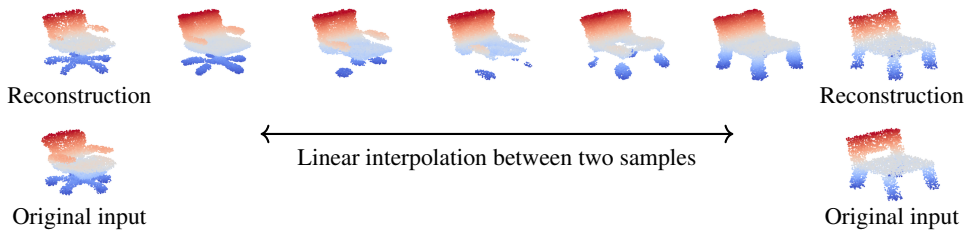}
    \caption{Linear interpolation in the latent space between two ShapeNet chair samples.
        The results exhibit smooth transitions in geometry, indicating that the latent space is continuous and semantically meaningful.}
    \label{fig:latent_interpolation}
\end{figure*}
To further examine the quality of the learned latent space, we conduct linear interpolations between the latent embeddings of two ShapeNet samples.
As shown in Fig.~\ref{fig:latent_interpolation}, the interpolated latent embeddings $\bm{z}$ yield reconstructions that not only achieve good fidelity to the original endpoints but also smoothly morph from one shape to the other.
The absence of abrupt artifacts suggests that the latent space learned by HyperINR is continuous and semantically meaningful, rather than memorizing isolated training instances.

% \subsection{sampling number}
% \begin{table}[t]
%     \caption{Effect of number of sampled queries on reconstruction/classification performance.}
%     \label{tab:sampling-number}
%     \centering
%     \begin{tabular}{lc}
%         \toprule
%         \textbf{Number of queries} & \textbf{Performance (\%)} \\
%         \midrule
%         1k                         & 90                        \\
%         10k                        & 97                        \\
%         60k                        & 98.18                     \\
%         \bottomrule
%     \end{tabular}
% \end{table}

\subsection{Ablation Studies}
\begin{table}[ht]
    \caption{Ablation on latent dimension $z$ for MNIST classification.
        All other components are fixed to the baseline configuration.}
    \label{tab:ablation_latent}
    \centering
    \begin{small}
        \begin{sc}

            \begin{tabular}{cc}
                \toprule
                Latent Dim. $z$ & Accuracy (\%)  \\
                \midrule
                10              & 81.01          \\
                \textbf{20}     & \textbf{98.13} \\
                64              & 97.44          \\
                128             & 96.24          \\
                256             & 91.40          \\
                \bottomrule
            \end{tabular}
        \end{sc}
    \end{small}
\end{table}

\begin{table}[ht]
    \caption{Ablation on MainNet width for MNIST classification.
        Latent dimension is fixed to $z=20$.}
    \label{tab:ablation_width}
    \centering
    \begin{small}
        \begin{sc}

            \begin{tabular}{cc}
                \toprule
                MainNet Width & Accuracy (\%)  \\
                \midrule
                32            & 97.24          \\
                \textbf{64}   & \textbf{98.13} \\
                128           & 97.42          \\
                256           & 94.94          \\
                \bottomrule
            \end{tabular}
        \end{sc}
    \end{small}
\end{table}
For ablation study, we adopt a unified baseline architecture (Table~\ref{tab:baseline_config} in Appendix~\ref{app:ablation}) and vary only a single factor at a time.

\paragraph{Latent dimension.}
A key hypothesis of our work is that INR weights reside on a low-dimensional manifold.
Table~\ref{tab:ablation_latent} evaluates classification accuracy as a function of latent dimension $z$, with all other components fixed.
Performance does not improve monotonically with dimension; instead, it peaks at an intrinsic dimension of 20, suggesting that higher dimensionality does not help find this low-dimensional manifold.
Notably, this contrasts with prior work such as inr2vec~\citep{luigi2023deep}, which assumes latent dimensions of 512 or 1024.
More reasonably, this suggests that the dataset MNIST lies close to a low-dimensional manifold with intrinsic dimension around 20.

\paragraph{MainNet width.}
We further ablate the width of the MainNet while fixing the latent dimension.
Results in Table~\ref{tab:ablation_width} show that a moderate width of 64 suffices to parameterize the learned functional manifold.
Wider networks yield diminishing or negative returns, indicating that performance is not driven by excessive MainNet capacity.
Larger widths may expand the weight space unnecessarily, increasing optimization complexity and making it harder to reliably discover the underlying low-dimensional structure.

\subsection{A Lightweight Setup and Matched Comparison Protocol}
As a theory-oriented study, we intentionally adopt a minimal experimental setup, using only the original loss functions and avoiding data augmentation, regularization, or auxiliary optimization techniques.
Due to fundamental architectural differences across prior works, it is difficult to evaluate all methods under an exactly matched parameterization or training pipeline.
Instead, we adopt a matched comparison protocol, in which all methods are evaluated under lightweight or tightly controlled settings, with comparable classifier capacity and without dataset enlargement or auxiliary optimization strategies.
Under this protocol, classification accuracy is interpreted as an indicator of semantic fidelity in the learned representations, rather than as a fully optimized benchmark metric.

The architectural complexity observed in many prior methods is largely a consequence of their modeling objectives.
In particular, several works focus on enforcing symmetry, equivariance, or invariance properties at the architectural level, which naturally requires highly expressive models and extensive data augmentation, including augmentation directly in the INR weight space.
For example, DWSNet~\citep{navon2023equivariant} relies on extensive data augmentation, while NFN~\citep{zhou2023permutation} employs multiple equivariant NF-layers with wide channel dimensions followed by large fully connected classifiers.
Similarly, inr2vec~\citep{luigi2023deep} employs a large encoder (four layers of 512, 512, 1024, and 1024 units) and a decoder with four hidden layers of 512 units each; their INR main network is a sizable MLP with four hidden layers of 512 units, and they report that reducing its size compromises reconstruction quality.
Inr2Array provides another representative example of this paradigm: on MNIST, it achieves strong reported performance by employing dataset enlargement via multiple independently initialized INRs per data instance as NFN does, a large transformer-based classifier, and extensive optimization, resulting in approximately $22$M parameters and $\sim27$ hours of training on a single NVIDIA A5000 GPU.
Such design choices are well aligned with its goal of enforcing invariance through architectural expressivity and augmented sampling.
In contrast, our approach targets a different objective: learning a compact, low-dimensional latent representation that captures the intrinsic semantic structure of the INR weight space itself.
This perspective enables a significantly more lightweight design, using a compact hypernetwork-based architecture with approximately $1$M parameters, most of which arise from the unavoidable linear mappings from latent embeddings to INR weights.
Despite this restrained setup, our joint learning framework converges within $30$ minutes on the same hardware and achieves competitive performance.
Detailed comparisons with every baseline are provided in Appendix~\ref{app:unifying_protocol}.

\section{Conclusion}
With the growing trend of treating weight as a new data modality, we present the HyperINR model as an instrument to ensure that semantic information from data can be reflected in weight representations.
Our framework employs a hypernetwork that maps a learnable low-dimensional latent space into the weight space of an Implicit Neural Representation (INR).
The learned latent embeddings exhibit natural clustering patterns, and classification experiments indicate that this representation effectively retains the semantics of the original data.
Finally, we use the Implicit Function Theorem (IFT) to outline the mapping between data and weight latent representations, offering a potential possibility for the future research in weight space learning.

\bibliographystyle{unsrt} 
\bibliography{ref}

@inproceedings{schurholt2024neural,
  title     = {Neural Network Weights as a New Data Modality},
  author    = {Konstantin Sch{\"u}rholt and Giorgos Bouritsas and Eliahu Horwitz and Derek Lim and Yoav Gelberg and Bo Zhao and Allan Zhou and Damian Borth and Stefanie Jegelka},
  booktitle = {ICLR 2025 Workshop Proposals},
  year      = {2024},
  url       = {https://openreview.net/forum?id=Bz6wEdobY7}
}

@inproceedings{luigi2023deep,
  title     = {Deep Learning on Implicit Neural Representations of Shapes},
  author    = {Luca De Luigi and Adriano Cardace and Riccardo Spezialetti and Pierluigi Zama Ramirez and Samuele Salti and Luigi di Stefano},
  booktitle = {The Eleventh International Conference on Learning Representations },
  year      = {2023},
  url       = {https://openreview.net/forum?id=OoOIW-3uadi}
}

@inproceedings{gielisse2025end,
  title={End-to-End Implicit Neural Representations for Classification},
  author={Gielisse, Alexander and van Gemert, Jan},
  booktitle={Proceedings of the Computer Vision and Pattern Recognition Conference},
  pages={18728--18737},
  year={2025}
}

@inproceedings{huang2025few,
  title={Few-shot Implicit Function Generation via Equivariance},
  author={Huang, Suizhi and Yang, Xingyi and Lu, Hongtao and Wang, Xinchao},
  booktitle={Proceedings of the Computer Vision and Pattern Recognition Conference},
  pages={16262--16272},
  year={2025}
}

@inproceedings{dupont2022data,
  title     = {From data to functa: Your data point is a function and you can treat it like one},
  author    = {Dupont, Emilien and Kim, Hyunjik and Eslami, S. M. Ali and Rezende, Danilo Jimenez and Rosenbaum, Dan},
  booktitle = {Proceedings of the 39th International Conference on Machine Learning},
  pages     = {5694--5725},
  year      = {2022},
  editor    = {Chaudhuri, Kamalika and Jegelka, Stefanie and Song, Le and Szepesvari, Csaba and Niu, Gang and Sabato, Sivan},
  volume    = {162},
  series    = {Proceedings of Machine Learning Research},
  month     = {17--23 Jul},
  publisher = {PMLR},
  url       = {https://proceedings.mlr.press/v162/dupont22a.html}
}

@article{zhou2023permutation,
  title={Permutation equivariant neural functionals},
  author={Zhou, Allan and Yang, Kaien and Burns, Kaylee and Cardace, Adriano and Jiang, Yiding and Sokota, Samuel and Kolter, J Zico and Finn, Chelsea},
  journal={Advances in neural information processing systems},
  volume={36},
  pages={24966--24992},
  year={2023}
}

@inproceedings{navon2023equivariant,
  title={Equivariant architectures for learning in deep weight spaces},
  author={Navon, Aviv and Shamsian, Aviv and Achituve, Idan and Fetaya, Ethan and Chechik, Gal and Maron, Haggai},
  booktitle={International Conference on Machine Learning},
  pages={25790--25816},
  year={2023},
  organization={PMLR}
}

@article{zhou2023neural,
  title={Neural functional transformers},
  author={Zhou, Allan and Yang, Kaien and Jiang, Yiding and Burns, Kaylee and Xu, Winnie and Sokota, Samuel and Kolter, J Zico and Finn, Chelsea},
  journal={Advances in neural information processing systems},
  volume={36},
  pages={77485--77502},
  year={2023}
}

@article{chang2015shapenet,
  title={Shapenet: An information-rich 3d model repository},
  author={Chang, Angel X and Funkhouser, Thomas and Guibas, Leonidas and Hanrahan, Pat and Huang, Qixing and Li, Zimo and Savarese, Silvio and Savva, Manolis and Song, Shuran and Su, Hao and others},
  journal={arXiv preprint arXiv:1512.03012},
  year={2015}
}

@article{qin2019pointdan,
  title={Pointdan: A multi-scale 3d domain adaption network for point cloud representation},
  author={Qin, Can and You, Haoxuan and Wang, Lichen and Kuo, C-C Jay and Fu, Yun},
  journal={Advances in Neural Information Processing Systems},
  volume={32},
  year={2019}
}

@article{ravi2020accelerating,
  title={Accelerating 3d deep learning with pytorch3d},
  author={Ravi, Nikhila and Reizenstein, Jeremy and Novotny, David and Gordon, Taylor and Lo, Wan-Yen and Johnson, Justin and Gkioxari, Georgia},
  journal={arXiv preprint arXiv:2007.08501},
  year={2020}
}

@article{sitzmann2020implicit,
  title={Implicit neural representations with periodic activation functions},
  author={Sitzmann, Vincent and Martel, Julien and Bergman, Alexander and Lindell, David and Wetzstein, Gordon},
  journal={Advances in neural information processing systems},
  volume={33},
  pages={7462--7473},
  year={2020}
}

@article{horwitz2025we,
  title={We Should Chart an Atlas of All the World's Models},
  author={Horwitz, Eliahu and Kurer, Nitzan and Kahana, Jonathan and Amar, Liel and Hoshen, Yedid},
  journal={arXiv preprint arXiv:2503.10633},
  year={2025}
}

@inproceedings{gal2016dropout,
  title={Dropout as a bayesian approximation: Representing model uncertainty in deep learning},
  author={Gal, Yarin and Ghahramani, Zoubin},
  booktitle={international conference on machine learning},
  pages={1050--1059},
  year={2016},
  organization={PMLR}
}

@inproceedings{finn2017model,
  title={Model-agnostic meta-learning for fast adaptation of deep networks},
  author={Finn, Chelsea and Abbeel, Pieter and Levine, Sergey},
  booktitle={International conference on machine learning},
  pages={1126--1135},
  year={2017},
  organization={PMLR}
}

@inproceedings{karras2019style,
  title={A style-based generator architecture for generative adversarial networks},
  author={Karras, Tero and Laine, Samuli and Aila, Timo},
  booktitle={Proceedings of the IEEE/CVF conference on computer vision and pattern recognition},
  pages={4401--4410},
  year={2019}
}

@inproceedings{ruiz2024hyperdreambooth,
  title={Hyperdreambooth: Hypernetworks for fast personalization of text-to-image models},
  author={Ruiz, Nataniel and Li, Yuanzhen and Jampani, Varun and Wei, Wei and Hou, Tingbo and Pritch, Yael and Wadhwa, Neal and Rubinstein, Michael and Aberman, Kfir},
  booktitle={Proceedings of the IEEE/CVF conference on computer vision and pattern recognition},
  pages={6527--6536},
  year={2024}
}

@inproceedings{phang2023hypertuning,
  title={Hypertuning: Toward adapting large language models without back-propagation},
  author={Phang, Jason and Mao, Yi and He, Pengcheng and Chen, Weizhu},
  booktitle={International Conference on Machine Learning},
  pages={27854--27875},
  year={2023},
  organization={pmlr}
}

@article{hornik1989multilayer,
  title={Multilayer feedforward networks are universal approximators},
  author={Hornik, Kurt and Stinchcombe, Maxwell and White, Halbert},
  journal={Neural networks},
  volume={2},
  number={5},
  pages={359--366},
  year={1989},
  publisher={Elsevier}
}

@InProceedings{pmlr-v119-spurek20a,
  title = 	 {Hypernetwork approach to generating point clouds},
  author =       {Spurek, Przemys{\l}aw and Winczowski, Sebastian and Tabor, Jacek and Zamorski, Maciej and Zieba, Maciej and Trzcinski, Tomasz},
  booktitle = 	 {Proceedings of the 37th International Conference on Machine Learning},
  pages = 	 {9099--9108},
  year = 	 {2020},
  editor = 	 {III, Hal Daumé and Singh, Aarti},
  volume = 	 {119},
  series = 	 {Proceedings of Machine Learning Research},
  month = 	 {13--18 Jul},
  publisher =    {PMLR},
  url = 	 {https://proceedings.mlr.press/v119/spurek20a.html}
}

@inproceedings{ha2016hypernetworks,
  title={Hypernetworks},
  author={Ha, David and Dai, Andrew and Le, Quoc V},
  booktitle={The Fifth International Conference on Learning Representations},
  year={2017}
}

@article{yang2024model,
  title={Model merging in llms, mllms, and beyond: Methods, theories, applications and opportunities},
  author={Yang, Enneng and Shen, Li and Guo, Guibing and Wang, Xingwei and Cao, Xiaochun and Zhang, Jie and Tao, Dacheng},
  journal={arXiv preprint arXiv:2408.07666},
  year={2024}
}

@inproceedings{park2019deepsdf,
  title={Deepsdf: Learning continuous signed distance functions for shape representation},
  author={Park, Jeong Joon and Florence, Peter and Straub, Julian and Newcombe, Richard and Lovegrove, Steven},
  booktitle={Proceedings of the IEEE/CVF conference on computer vision and pattern recognition},
  pages={165--174},
  year={2019}
}

@inproceedings{kidger2020universal,
  title={Universal approximation with deep narrow networks},
  author={Kidger, Patrick and Lyons, Terry},
  booktitle={Conference on learning theory},
  pages={2306--2327},
  year={2020},
  organization={PMLR}
}

@article{schurholt2022hyper,
  title={Hyper-representations as generative models: Sampling unseen neural network weights},
  author={Sch{\"u}rholt, Konstantin and Knyazev, Boris and Gir{\'o}-i-Nieto, Xavier and Borth, Damian},
  journal={Advances in Neural Information Processing Systems},
  volume={35},
  pages={27906--27920},
  year={2022}
}

@inproceedings{gordon2024d,
  title={D'OH: Decoder-Only Random Hypernetworks for Implicit Neural Representations},
  author={Gordon, Cameron and MacDonald, Lachlan E and Saratchandran, Hemanth and Lucey, Simon},
  booktitle={Proceedings of the Asian Conference on Computer Vision},
  pages={2507--2526},
  year={2024}
}

@article{yann2010mnist,
  title={MNIST handwritten digit database},
  author={Yann, LeCun},
  journal={ATT Labs.},
  year={2010}
}

@article{xiao2017fashion,
  title={Fashion-mnist: a novel image dataset for benchmarking machine learning algorithms},
  author={Xiao, Han and Rasul, Kashif and Vollgraf, Roland},
  journal={arXiv preprint arXiv:1708.07747},
  year={2017}
}

@inproceedings{wu20153d,
  title={3d shapenets: A deep representation for volumetric shapes},
  author={Wu, Zhirong and Song, Shuran and Khosla, Aditya and Yu, Fisher and Zhang, Linguang and Tang, Xiaoou and Xiao, Jianxiong},
  booktitle={Proceedings of the IEEE conference on computer vision and pattern recognition},
  pages={1912--1920},
  year={2015}
}

@inproceedings{mullen2010signing,
  title={Signing the unsigned: Robust surface reconstruction from raw pointsets},
  author={Mullen, Patrick and De Goes, Fernando and Desbrun, Mathieu and Cohen-Steiner, David and Alliez, Pierre},
  booktitle={Computer Graphics Forum},
  volume={29},
  pages={1733--1741},
  year={2010},
  organization={Wiley Online Library}
}

@book{jolliffe2002pca, 
  title={Principal Component Analysis},
  author={I. T. Jolliffe},
  year={2002},
  edition = {2nd},
  publisher={Springer}
}

@inproceedings{houlsby2019parameter,
  title={Parameter-efficient transfer learning for NLP},
  author={Houlsby, Neil and Giurgiu, Andrei and Jastrzebski, Stanislaw and Morrone, Bruna and De Laroussilhe, Quentin and Gesmundo, Andrea and Attariyan, Mona and Gelly, Sylvain},
  booktitle={International conference on machine learning},
  pages={2790--2799},
  year={2019},
  organization={PMLR}
}

@inproceedings{gatys2016image,
  title={Image style transfer using convolutional neural networks},
  author={Gatys, Leon A and Ecker, Alexander S and Bethge, Matthias},
  booktitle={Proceedings of the IEEE conference on computer vision and pattern recognition},
  pages={2414--2423},
  year={2016}
}

@inproceedings{gu2023generalizable,
  title={Generalizable neural fields as partially observed neural processes},
  author={Gu, Jeffrey and Wang, Kuan-Chieh and Yeung, Serena},
  booktitle={Proceedings of the IEEE/CVF International Conference on Computer Vision},
  pages={5330--5339},
  year={2023}
}

@article{sitzmann2019scene,
  title={Scene representation networks: Continuous 3d-structure-aware neural scene representations},
  author={Sitzmann, Vincent and Zollh{\"o}fer, Michael and Wetzstein, Gordon},
  journal={Advances in neural information processing systems},
  volume={32},
  year={2019}
}

@inproceedings{
kahana2025deep,
title={Deep Linear Probe Generators for Weight Space Learning},
author={Jonathan Kahana and Eliahu Horwitz and Imri Shuval and Yedid Hoshen},
booktitle={The Thirteenth International Conference on Learning Representations},
year={2025},
url={https://openreview.net/forum?id=XoYdD3m0mv}
}

@inproceedings{navon2024equivariant,
author = {Navon, Aviv and Shamsian, Aviv and Fetaya, Ethan and Chechik, Gal and Dym, Nadav and Maron, Haggai},
title = {Equivariant deep weight space alignment},
year = {2024},
publisher = {JMLR.org},
booktitle = {Proceedings of the 41st International Conference on Machine Learning},
articleno = {1518},
numpages = {20},
location = {Vienna, Austria},
series = {ICML'24}
}

@article{gu2025foundation,
  title={Foundation models secretly understand neural network weights: Enhancing hypernetwork architectures with foundation models},
  author={Gu, Jeffrey and Yeung-Levy, Serena},
  journal={The Thirteenth International Conference on Learning Representations},
  year={2025}
}

@inproceedings{chen2022transformers,
  title={Transformers as meta-learners for implicit neural representations},
  author={Chen, Yinbo and Wang, Xiaolong},
  booktitle={European Conference on Computer Vision},
  pages={170--187},
  year={2022},
  organization={Springer}
}

@inproceedings{kim2023generalizable,
  title={Generalizable implicit neural representations via instance pattern composers},
  author={Kim, Chiheon and Lee, Doyup and Kim, Saehoon and Cho, Minsu and Han, Wook-Shin},
  booktitle={Proceedings of the IEEE/CVF Conference on Computer Vision and Pattern Recognition},
  pages={11808--11817},
  year={2023}
}

@book{krantz2002implicit,
  title={The implicit function theorem: history, theory, and applications},
  author={Krantz, Steven George and Parks, Harold R},
  year={2002},
  publisher={Springer Science \& Business Media}
}

%%%%%%%%%%%%%%%%%%%%%%%%%%%%%%%%%%%%%%%%%%%%%%%%%%%%%%%%%%%%%%%%%%%%%%%%%%%%%%%
%%%%%%%%%%%%%%%%%%%%%%%%%%%%%%%%%%%%%%%%%%%%%%%%%%%%%%%%%%%%%%%%%%%%%%%%%%%%%%%
% APPENDIX
%%%%%%%%%%%%%%%%%%%%%%%%%%%%%%%%%%%%%%%%%%%%%%%%%%%%%%%%%%%%%%%%%%%%%%%%%%%%%%%
%%%%%%%%%%%%%%%%%%%%%%%%%%%%%%%%%%%%%%%%%%%%%%%%%%%%%%%%%%%%%%%%%%%%%%%%%%%%%%%
\newpage
\appendix
\onecolumn

\section{Dataset Details}
\label{sec:appendix-dataset}
We report dataset statistics including the number of samples, classes, and train/test splits.
The 2D datasets can be directly used for INR reconstruction, whereas the 3D meshes require conversion to a coordinate-based representation~\citep{mullen2010signing}.
For 3D point clouds, we uniformly sample 10K queries and compute unsigned distance function (UDF) values by building a KDTree on the fitted point cloud.
The nearest-neighbor distance is taken as the UDF value, using the PyTorch3D implementation~\citep{ravi2020accelerating}.
No augmentation, noise, normalization, or rescaling is applied.
Prior works often prepare up to 500K query points near surfaces and then sample different subsets of 10K points during training as data augmentation to improve INR robustness~\citep{luigi2023deep}.
In contrast, we just uniformly sample 10K queries and obtain the ground truth UDF values via a KDTree~\citep{ravi2020accelerating}.
This simple setup is adequate for validating our concept.
\begin{table}[ht]
    \caption{
        Dataset statistics.
    }
    \label{tab:dataset-stats}
    \begin{center}

        \begin{small}
            \begin{sc}
                \begin{tabular}{lccc}
                    \toprule
                    \textbf{Dataset} & \textbf{Classes} & \textbf{Train} & \textbf{Test} \\
                    \midrule
                    MNIST            & 10               & 60,000         & 10,000        \\
                    FashionMNIST     & 10               & 60,000         & 10,000        \\
                    ModelNet40       & 40               & 9,843          & 2,468         \\
                    ShapeNet10       & 10               & 20,409         & 5,103         \\
                    ScanNet10        & 10               & 6,110          & 1,769         \\
                    \bottomrule
                \end{tabular}
            \end{sc}
        \end{small}

    \end{center}
\end{table}

\section{Empirical verification of IFT conditions: Hessian rank}
\label{app:hess}

To verify the full rank condition required by our IFT analysis, we numerically computed the Hessian matrices for the full training set and test set of the five used datasets. We analyzed the spectral properties of these Hessians to determine their smallest singular values ($\sigma_{\min}$) and highest condition numbers ($\kappa$), as well as their distributions.

% Figure placement
\begin{figure}[ht]
    \centering
    \includegraphics[width=0.48\textwidth]{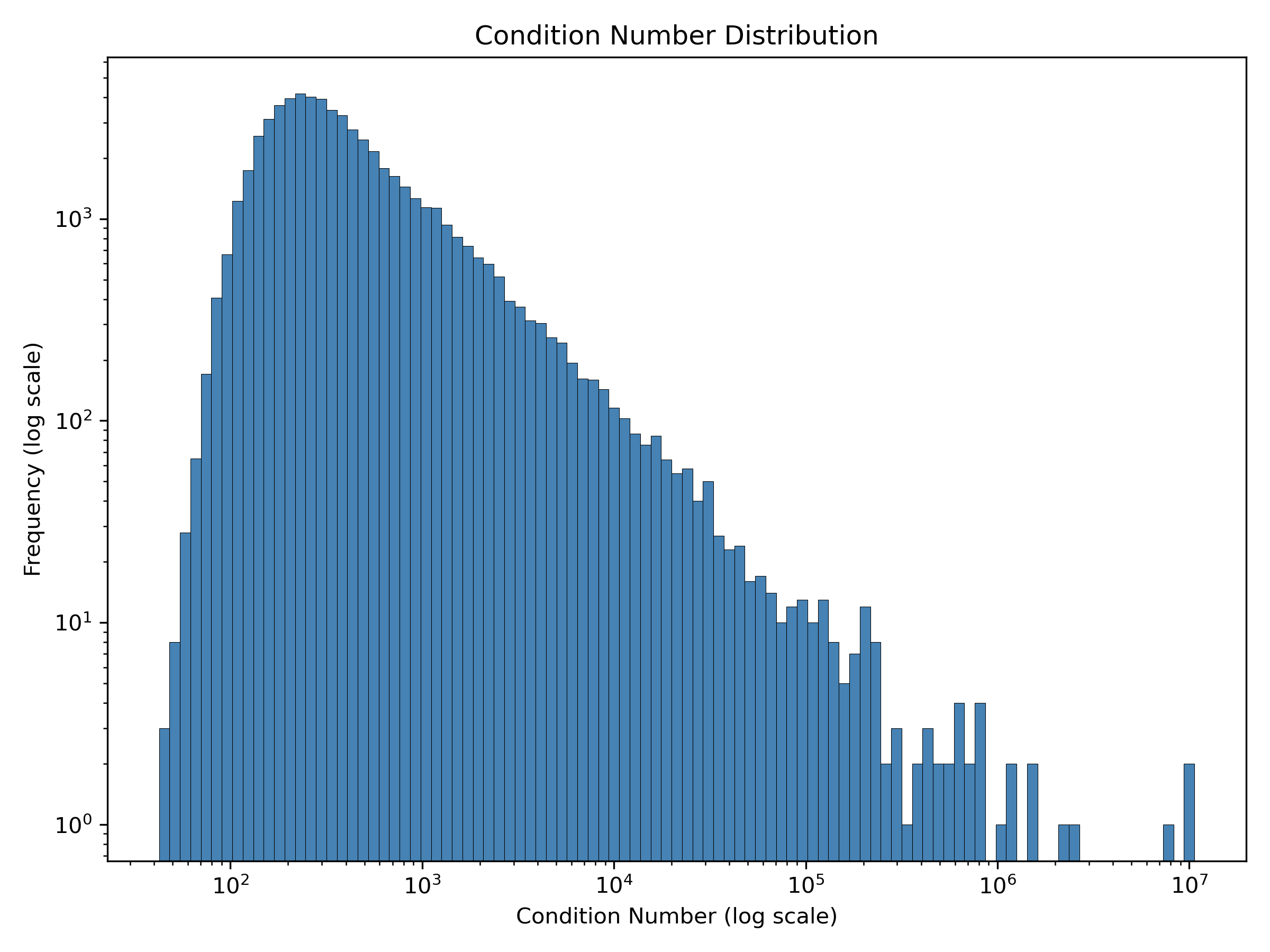}
    \hfill
    \includegraphics[width=0.48\textwidth]{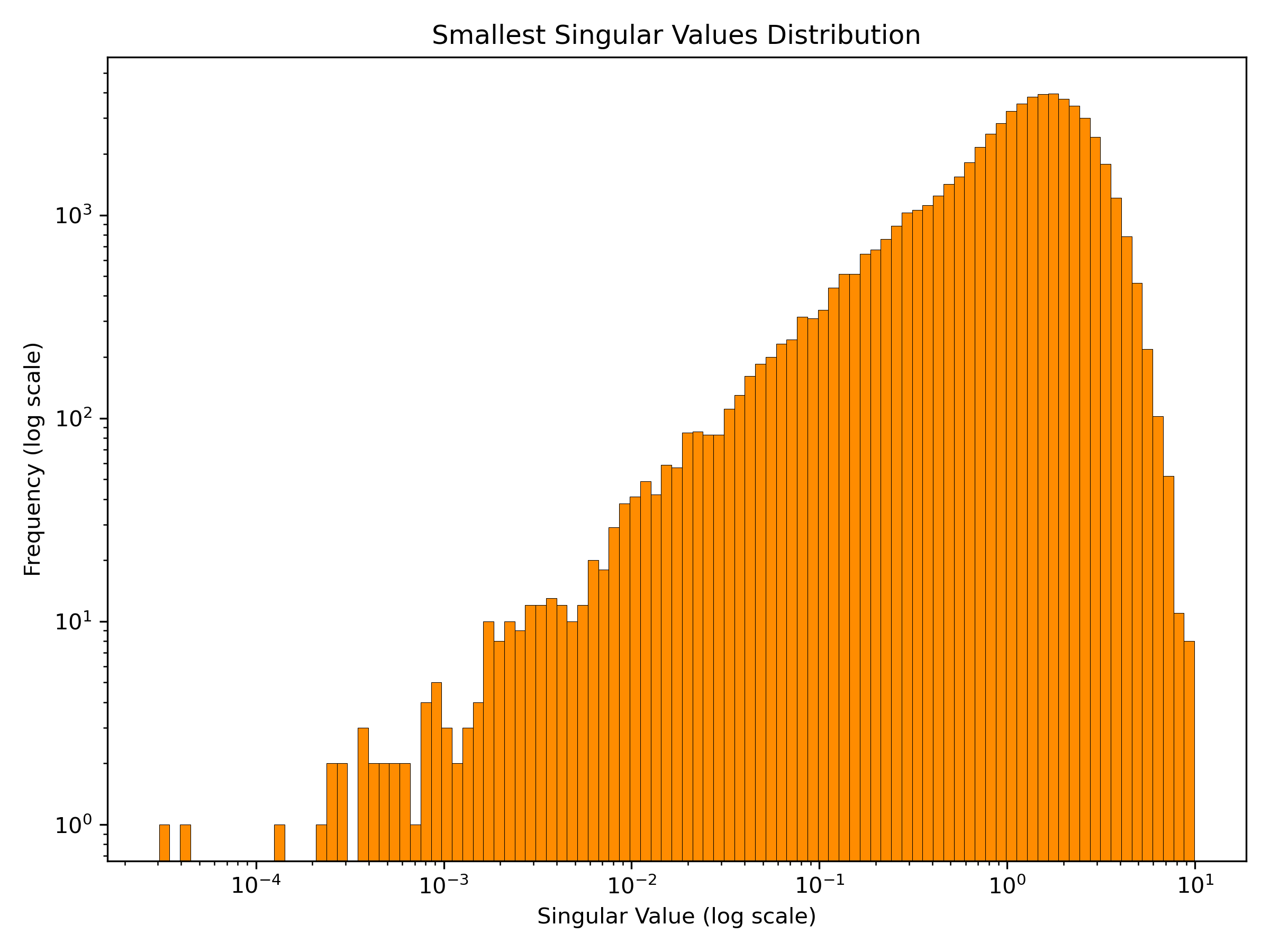}
    
    \caption{\textbf{Hessian analysis on MNIST training data (log-log scale).}The distribution of condition numbers (\textbf{left}) shows that the probability mass is concentrated in the well-conditioned range ($10^2 - 10^3$).High condition numbers ($\kappa > 10^5$) appear only as rare, isolated outliers in the long tail. Correspondingly, the smallest singular values (\textbf{right}) remain strictly above the float32 machine epsilon ($\sim 1.2 \cdot 10^{-7}$), empirically confirming that the Hessians are full-rank and invertible.}

    \label{fig:mnist_hessian_analysis}
\end{figure}
%:1.Distribution of Hessian condition numbers on MNIST.}} \textcolor{red}{The histogram (log-log scale) reveals that the vast majority of Hessians are well-conditioned, with the probability mass concentrated between $10^2$ and $10^3$. High condition numbers ($\kappa > 10^5$) appear only as rare, isolated outliers in the long tail.
% 2.The histogram (log-log scale) reveals that all Hessians have full rank and thus are invertible, since all are well above the float32 machine epsilon $(\sim 1.2 \cdot 10^{-7})$
%3. The histogram (log-log scale) reveals that the vast majority of Hessians are well-conditioned, with the probability mass concentrated between $10^2$ and $10^3$. High condition numbers ($\kappa > 10^5$) appear only as rare, isolated outliers in the long tail.
% 4.Distribution of the smallest singular values on FashionMNIST.}} \textcolor{red}{The histogram (log-log scale) reveals that all Hessians have full rank and thus are invertible, since all are well above the float32 machine epsilon $(\sim 1.2 \cdot 10^{-7})$

\begin{figure}[ht]
    \centering
    \includegraphics[width=0.48\textwidth]{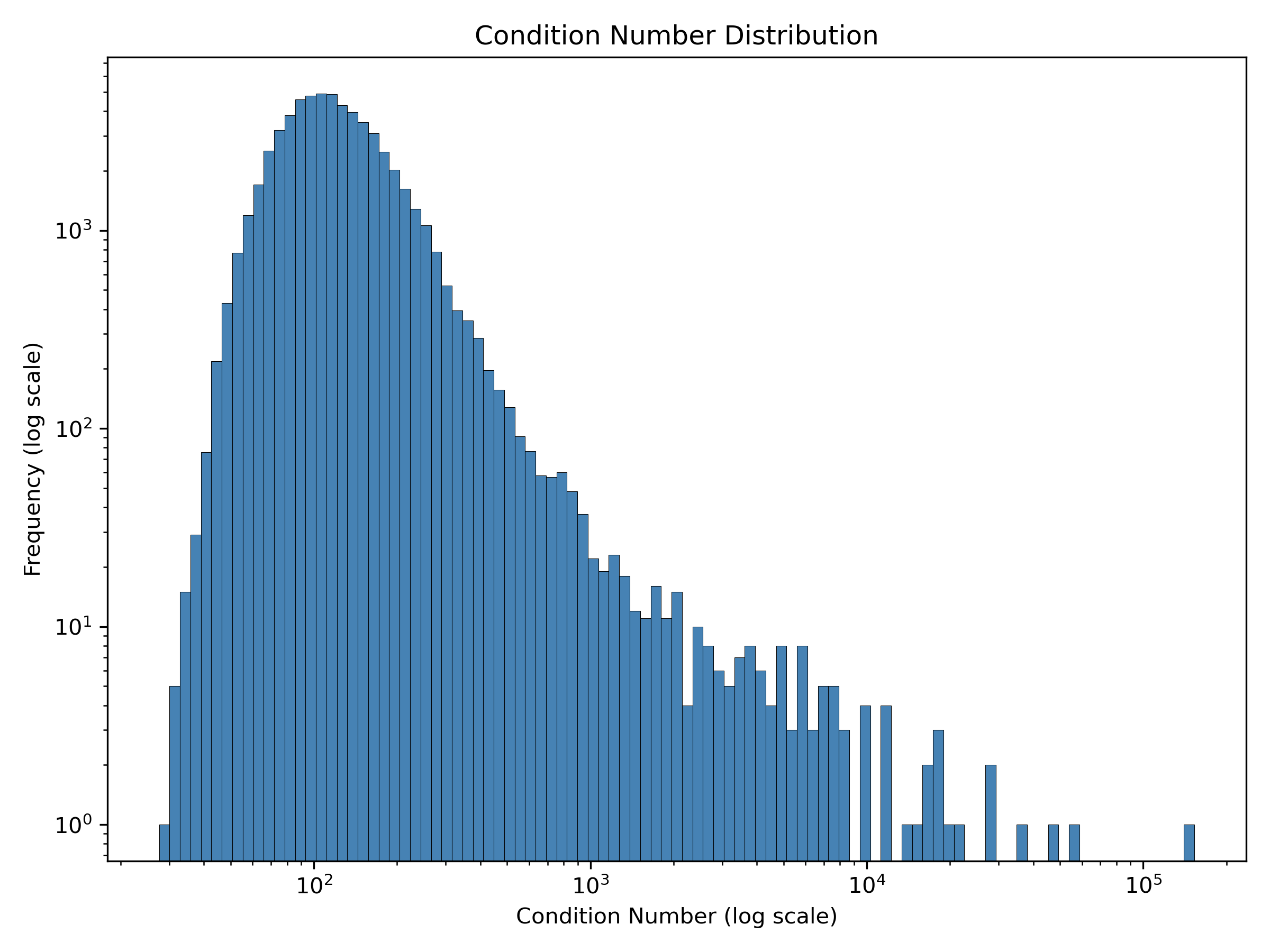}
    \hfill
    \includegraphics[width=0.48\textwidth]{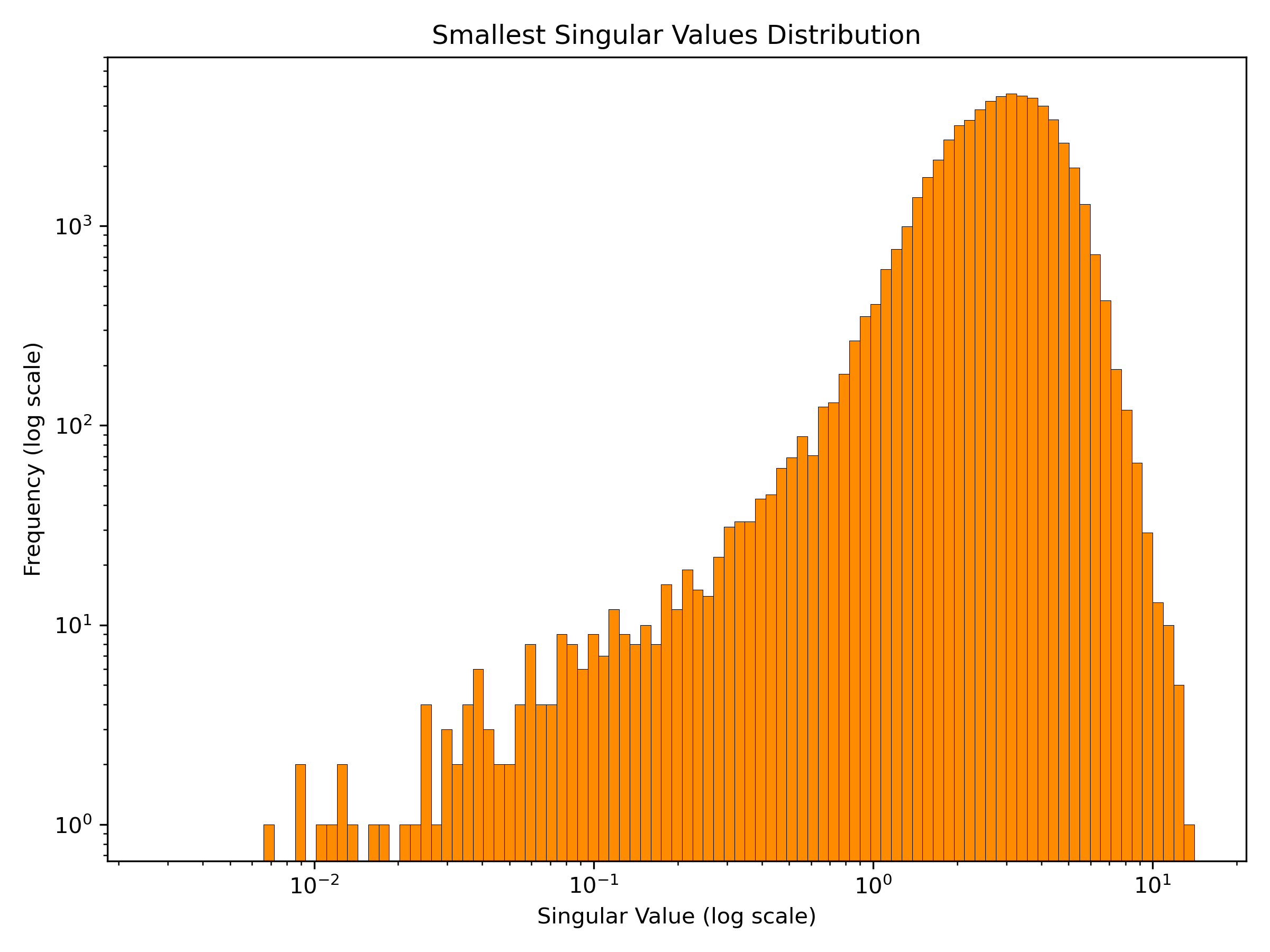}
    \caption{\textbf{Hessian analysis on FashionMNIST training data (log-log scale).}The distribution of condition numbers (\textbf{left}) shows that the probability mass is concentrated in the well-conditioned range ($10^2 - 10^3$), with negligible outliers. Correspondingly, the smallest singular values (\textbf{right}) remain strictly above the float32 machine epsilon ($\sim 1.2 \cdot 10^{-7}$), empirically confirming that the Hessians are full-rank and invertible.}
    \label{fig:fmnist_hessian_analysis}
\end{figure}

\paragraph{Numerical non-singularity} A critical requirement for the IFT is that the Jacobian (and by extension the Hessian in our implicit function definition) is non-singular. As shown in Table~\ref{tab:mnist_sing}, the smallest singular values for MNIST are approximately $3.05 \cdot 10^{-5}$ (training set) and $2.53 \cdot 10^{-5}$ (test set).
For FashionMNIST (Table~\ref{tab:fmnist_sing}), they are $2.79 \cdot 10^{-3}$ (training set) and $3.95 \cdot 10^{-5}$ (test set). Crucially, these values are orders of magnitude above the float32 machine epsilon ($\epsilon \approx 1.2 \cdot 10^{-7}$). This confirms that the Hessians are numerically full rank across both 2D datasets. ScanNet10 has a similar smallest singular value distribution, as depicted in Table~\ref{tab:scannet_sing}.
The two remaining 3D datasets contain more outliers and have a very small percentage of samples approaching non singularity, as evident by the distribution of their smallest singular values: $0.19\%$ of ModelNet40 training set and $0.04\%$ of ModelNet40 test set samples lie below $10^{-6}$ in Table~\ref{tab:modelnet40_sing}, and $0.27\%$ of ShapeNet10 training set and $0.25\%$ of ShapeNet test samples lie below $10^{-6}$ in Table~\ref{tab:shapenet10_sing}.

\paragraph{Condition number} We further analyze the stability of the mapping $g$ via the condition number distribution, detailed in Tables~\ref{tab:mnist_cond} and \ref{tab:fmnist_cond} for the 2D datasets and in Tables~\ref{tab:modelnet40_cond}, \ref{tab:shapenet10_cond}  and \ref{tab:scannet_cond} for the 3D datasets.
As illustrated in Fig.~\ref{fig:mnist_hessian_analysis} and \ref{fig:fmnist_hessian_analysis}, the distribution of condition numbers for both 2D datasets exhibits a clear concentration of mass in the well-conditioned regime (peaking between $10^2$ and $10^3$).
While the theoretical maximum condition numbers reach $\sim 10^7$, the log-scale frequency axis highlights that these are extreme outliers; the curve decays rapidly, rendering the ill-conditioned tail statistically negligible.
For instance, in FashionMNIST, less than $0.05\%$ of training samples exceed a condition number of $10^4$.
The 3D datasets exhibit similar distributions (Tables~\ref{tab:modelnet40_cond}, \ref{tab:shapenet10_cond}, \ref{tab:scannet_cond}).
These empirical distributions align closely with our theoretical expectations, suggesting that the local mapping described by the IFT is largely consistent with observations in the weight space for our datasets, with only rare, localized instabilities.

% ------------------------------------------------------------------
% MNIST TABLES
% ------------------------------------------------------------------

\begin{table}[t]
    \caption{Cumulative distribution of Hessian condition numbers on MNIST}
    \label{tab:mnist_cond}
    \begin{center}

        \begin{small}
            \begin{sc}
                \begin{tabular}{lll}
                    \toprule
                    \multicolumn{1}{c}{ Threshold ($\tau$)}          &
                    \multicolumn{1}{c}{ \% Samples $> \tau$ (Train)} &
                    \multicolumn{1}{c}{ \% Samples $> \tau$ (Test)}                                                \\
                    \midrule
                    $>10^{1}$                                        & $100.000$            & $100.000$            \\
                    $>10^{2}$                                        & $98.002$             & $98.710$             \\
                    $>10^{3}$                                        & $16.420$             & $20.180$             \\
                    $>10^{4}$                                        & $1.552$              & $2.170$              \\
                    $>10^{5}$                                        & $0.168$              & $0.290$              \\
                    $>10^{6}$                                        & $0.015$              & $0.050$              \\
                    $>10^{7}$                                        & $0.003$              & $0.010$              \\
                    $>10^{8}$                                        & $0.000$              & $0.000$              \\
                    \midrule
                    \textbf{Max. $\kappa$}                           & $1.065 \cdot 10^{7}$ & $1.227 \cdot 10^{7}$ \\
                    \bottomrule
                \end{tabular}
            \end{sc}
        \end{small}

    \end{center}
\end{table}

\begin{table}[t]
    \caption{Cumulative distribution of Hessian smallest singular values on MNIST}
    \label{tab:mnist_sing}
    \begin{center}

        \begin{small}
            \begin{sc}
                \begin{tabular}{lll}
                    \toprule
                    \multicolumn{1}{c}{ Threshold ($\tau$)}          &
                    \multicolumn{1}{c}{ \% Samples $< \tau$ (Train)} &
                    \multicolumn{1}{c}{ \% Samples $< \tau$ (Test)}                                                  \\
                    \midrule
                    $<10^{2}$                                        & $100.000$             & $100.000$             \\
                    $<10^{1}$                                        & $100.000$             & $100.000$             \\
                    $<10^{0}$                                        & $40.888$              & $47.230$              \\
                    $<10^{-1}$                                       & $4.617$               & $6.280$               \\
                    $<10^{-2}$                                       & $0.435$               & $0.740$               \\
                    $<10^{-3}$                                       & $0.050$               & $0.120$               \\
                    $<10^{-4}$                                       & $0.005$               & $0.010$               \\
                    $<10^{-5}$                                       & $0.000$               & $0.000$               \\
                    $<10^{-6}$                                       & $0.000$               & $0.000$               \\
                    \midrule
                    \textbf{Min. $\sigma_{\min}$}                    & $3.053 \cdot 10^{-5}$ & $2.534 \cdot 10^{-5}$ \\
                    \bottomrule
                \end{tabular}
            \end{sc}
        \end{small}

    \end{center}
\end{table}

\begin{table}[t]
    \caption{Cumulative distribution of Hessian condition numbers on FashionMNIST}
    \label{tab:fmnist_cond}
    \begin{center}

        \begin{small}
            \begin{sc}
                \begin{tabular}{lll}
                    \toprule
                    \multicolumn{1}{c}{ Threshold ($\tau$)}          &
                    \multicolumn{1}{c}{ \% Samples $> \tau$ (Train)} &
                    \multicolumn{1}{c}{ \% Samples $> \tau$ (Test)}                                                \\
                    \midrule
                    $>10^{1}$                                        & $100.000$            & $100.000$            \\
                    $>10^{2}$                                        & $62.672$             & $66.670$             \\
                    $>10^{3}$                                        & $0.427$              & $0.850$              \\
                    $>10^{4}$                                        & $0.035$              & $0.060$              \\
                    $>10^{5}$                                        & $0.002$              & $0.030$              \\
                    $>10^{6}$                                        & $0.000$              & $0.010$              \\
                    $>10^{7}$                                        & $0.000$              & $0.000$              \\
                    $>10^{8}$                                        & $0.000$              & $0.000$              \\
                    \midrule
                    \textbf{Max. $\kappa$}                           & $1.532 \cdot 10^{5}$ & $9.257 \cdot 10^{6}$ \\
                    \bottomrule
                \end{tabular}
            \end{sc}
        \end{small}

    \end{center}
\end{table}

\begin{table}[t]
    \caption{Cumulative distribution of Hessian smallest singular values on FashionMNIST}
    \label{tab:fmnist_sing}
    \begin{center}

        \begin{small}
            \begin{sc}
                \begin{tabular}{lll}
                    \toprule
                    \multicolumn{1}{c}{ Threshold ($\tau$)}          &
                    \multicolumn{1}{c}{ \% Samples $< \tau$ (Train)} &
                    \multicolumn{1}{c}{ \% Samples $< \tau$ (Test)}                                                  \\
                    \midrule
                    $<10^{2}$                                        & $100.000$             & $100.000$             \\
                    $<10^{1}$                                        & $99.950$              & $99.980$              \\
                    $<10^{0}$                                        & $3.102$               & $5.900$               \\
                    $<10^{-1}$                                       & $0.148$               & $0.270$               \\
                    $<10^{-2}$                                       & $0.007$               & $0.030$               \\
                    $<10^{-3}$                                       & $0.000$               & $0.020$               \\
                    $<10^{-4}$                                       & $0.000$               & $0.010$               \\
                    $<10^{-5}$                                       & $0.000$               & $0.000$               \\
                    $<10^{-6}$                                       & $0.000$               & $0.000$               \\
                    \midrule
                    \textbf{Min. $\sigma_{\min}$}                    & $2.789 \cdot 10^{-3}$ & $3.950 \cdot 10^{-5}$ \\
                    \bottomrule
                \end{tabular}
            \end{sc}
        \end{small}

    \end{center}
\end{table}

% ------------------------ ModelNet40 Hessian Tables ------------------------
\begin{table}[ht]
    \caption{Cumulative distribution of Hessian condition numbers on ModelNet40}
    \label{tab:modelnet40_cond}
    \begin{center}
        \begin{small}
            \begin{sc}
                \begin{tabular}{lll}
                    \toprule
                    \multicolumn{1}{c}{ Threshold ($\tau$)} & \multicolumn{1}{c}{ \% Samples $> \tau$ (Train)} & \multicolumn{1}{c}{ \% Samples $> \tau$ (Test)} \\
                    \midrule
                    $>10^{0}$                               & $100.00$                                         & $100.00$                                        \\
                    $>10^{1}$                               & $100.00$                                         & $100.00$                                        \\
                    $>10^{2}$                               & $94.23$                                          & $97.29$                                         \\
                    $>10^{3}$                               & $16.91$                                          & $6.60$                                          \\
                    $>10^{4}$                               & $1.78$                                           & $0.12$                                          \\
                    $>10^{5}$                               & $0.18$                                           & $0.04$                                          \\
                    $>10^{6}$                               & $0.03$                                           & $0.04$                                          \\
                    $>10^{7}$                               & $0.00$                                           & $0.00$                                          \\
                    $>10^{8}$                               & $0.00$                                           & $0.00$                                          \\
                    \midrule
                    \textbf{Max. $\kappa$}                  & $2.640264 \cdot 10^{6}$                          & $7.737528 \cdot 10^{6}$                         \\
                    \bottomrule
                \end{tabular}
            \end{sc}
        \end{small}
    \end{center}
\end{table}

\begin{table}[ht]
    \caption{Cumulative distribution of Hessian smallest singular values on ModelNet40}
    \label{tab:modelnet40_sing}
    \begin{center}
        \begin{small}
            \begin{sc}
                \begin{tabular}{lll}
                    \toprule
                    \multicolumn{1}{c}{ Threshold ($\tau$)} & \multicolumn{1}{c}{ \% Samples $< \tau$ (Train)} & \multicolumn{1}{c}{ \% Samples $< \tau$ (Test)} \\
                    \midrule
                    $<10^{-6}$                              & $0.19$                                           & $0.04$                                          \\
                    $<10^{-5}$                              & $1.71$                                           & $0.12$                                          \\
                    $<10^{-4}$                              & $16.20$                                          & $5.79$                                          \\
                    $<10^{-3}$                              & $92.47$                                          & $96.68$                                         \\
                    $<10^{-2}$                              & $100.00$                                         & $100.00$                                        \\
                    $<10^{-1}$                              & $100.00$                                         & $100.00$                                        \\
                    $<10^{0}$                               & $100.00$                                         & $100.00$                                        \\
                    \midrule
                    \textbf{Min. $\sigma_{\min}$}           & $4.40 \cdot 10^{-8}$                             & $1.72 \cdot 10^{-8}$                            \\
                    \bottomrule
                \end{tabular}
            \end{sc}
        \end{small}
    \end{center}
\end{table}

% ---------------------------------------------------------------------------
% ------------------------ ShapeNet10 Hessian Tables ------------------------
\begin{table}[ht]
    \caption{Cumulative distribution of Hessian condition numbers on ShapeNet10}
    \label{tab:shapenet10_cond}
    \begin{center}
        \begin{small}
            \begin{sc}
                \begin{tabular}{lll}
                    \toprule
                    \multicolumn{1}{c}{ Threshold ($\tau$)} & \multicolumn{1}{c}{ \% Samples $> \tau$ (Train)} & \multicolumn{1}{c}{ \% Samples $> \tau$ (Test)} \\
                    \midrule
                    $>10^{0}$                               & $100.00$                                         & $100.00$                                        \\
                    $>10^{1}$                               & $100.00$                                         & $100.00$                                        \\
                    $>10^{2}$                               & $97.55$                                          & $99.69$                                         \\
                    $>10^{3}$                               & $40.94$                                          & $58.20$                                         \\
                    $>10^{4}$                               & $2.69$                                           & $1.20$                                          \\
                    $>10^{5}$                               & $0.27$                                           & $0.25$                                          \\
                    $>10^{6}$                               & $0.03$                                           & $0.04$                                          \\
                    $>10^{7}$                               & $0.00$                                           & $0.00$                                          \\
                    $>10^{8}$                               & $0.00$                                           & $0.00$                                          \\
                    \midrule
                    \textbf{Max. $\kappa$}                  & $1.2486522 \cdot 10^{7}$                         & $1.245944 \cdot 10^{6}$                         \\
                    \bottomrule
                \end{tabular}
            \end{sc}
        \end{small}
    \end{center}
\end{table}

\begin{table}[ht]
    \caption{Cumulative distribution of Hessian smallest singular values on ShapeNet10}
    \label{tab:shapenet10_sing}
    \begin{center}
        \begin{small}
            \begin{sc}
                \begin{tabular}{lll}
                    \toprule
                    \multicolumn{1}{c}{ Threshold ($\tau$)} & \multicolumn{1}{c}{ \% Samples $< \tau$ (Train)} & \multicolumn{1}{c}{ \% Samples $< \tau$ (Test)} \\
                    \midrule
                    $<10^{-6}$                              & $0.27$                                           & $0.25$                                          \\
                    $<10^{-5}$                              & $2.65$                                           & $1.06$                                          \\
                    $<10^{-4}$                              & $38.56$                                          & $55.40$                                         \\
                    $<10^{-3}$                              & $94.87$                                          & $98.96$                                         \\
                    $<10^{-2}$                              & $100.00$                                         & $100.00$                                        \\
                    $<10^{-1}$                              & $100.00$                                         & $100.00$                                        \\
                    $<10^{0}$                               & $100.00$                                         & $100.00$                                        \\
                    \midrule
                    \textbf{Min. $\sigma_{\min}$}           & $8.30 \cdot 10^{-9}$                             & $8.05 \cdot 10^{-8}$                            \\
                    \bottomrule
                \end{tabular}
            \end{sc}
        \end{small}
    \end{center}
\end{table}

% ---------------------------------------------------------------------------

% ------------------------ ScanNet10 Hessian Tables -------------------------
\begin{table}[ht]
    \caption{Cumulative distribution of Hessian condition numbers on ScanNet10}
    \label{tab:scannet_cond}
    \begin{center}
        \begin{small}
            \begin{sc}
                \begin{tabular}{lll}
                    \toprule
                    \multicolumn{1}{c}{ Threshold ($\tau$)} & \multicolumn{1}{c}{ \% Samples $> \tau$ (Train)} & \multicolumn{1}{c}{ \% Samples $> \tau$ (Test)} \\
                    \midrule
                    $>10^{0}$                               & $100.00$                                         & $100.00$                                        \\
                    $>10^{1}$                               & $100.00$                                         & $100.00$                                        \\
                    $>10^{2}$                               & $64.70$                                          & $81.57$                                         \\
                    $>10^{3}$                               & $0.59$                                           & $0.40$                                          \\
                    $>10^{4}$                               & $0.02$                                           & $0.00$                                          \\
                    $>10^{5}$                               & $0.00$                                           & $0.00$                                          \\
                    $>10^{6}$                               & $0.00$                                           & $0.00$                                          \\
                    $>10^{7}$                               & $0.00$                                           & $0.00$                                          \\
                    \midrule
                    \textbf{Max. $\kappa$}                  & $2.504012 \cdot 10^{4}$                          & $5.67563 \cdot 10^{3}$                          \\
                    \bottomrule
                \end{tabular}
            \end{sc}
        \end{small}
    \end{center}
\end{table}

\begin{table}[ht]
    \caption{Cumulative distribution of Hessian smallest singular values on ScanNet10}
    \label{tab:scannet_sing}
    \begin{center}
        \begin{small}
            \begin{sc}
                \begin{tabular}{lll}
                    \toprule
                    \multicolumn{1}{c}{ Threshold ($\tau$)} & \multicolumn{1}{c}{ \% Samples $< \tau$ (Train)} & \multicolumn{1}{c}{ \% Samples $< \tau$ (Test)} \\
                    \midrule
                    $<10^{-6}$                              & $0.00$                                           & $0.00$                                          \\
                    $<10^{-5}$                              & $0.02$                                           & $0.00$                                          \\
                    $<10^{-4}$                              & $0.23$                                           & $0.23$                                          \\
                    $<10^{-3}$                              & $16.56$                                          & $26.96$                                         \\
                    $<10^{-2}$                              & $100.00$                                         & $100.00$                                        \\
                    $<10^{-1}$                              & $100.00$                                         & $100.00$                                        \\
                    $<10^{0}$                               & $100.00$                                         & $100.00$                                        \\
                    \midrule
                    \textbf{Min. $\sigma_{\min}$}           & $9.05 \cdot 10^{-6}$                             & $1.76 \cdot 10^{-5}$                            \\
                    \bottomrule
                \end{tabular}
            \end{sc}
        \end{small}
    \end{center}
\end{table}

% ---------------------------------------------------------------------------

\section{Network Architecture and Training}
\label{sec:appendix-architecture}
Table~\ref{tab:architecture-details} summarizes the architectures and training setups used across datasets.
We deliberately adopt small and simple networks, keeping the design consistent between 2D and 3D tasks.
The latent dimension $z$ is set to either 20 or 32, and all models are trained for 500 epochs in the first step of phase A.
The scaling factor for spatial frequency $\omega_0$ used in main net of SIREN architecture~\citep{sitzmann2020implicit} is 30 for all the data.
The paramenter for network stands for the hidden layer size.
E.g., ``HNet [256,256]'' means there are two hidden layers with the size of 256.

\begin{table}[ht]
    \caption{Architectures and training settings}
    \label{tab:architecture-details}
    \begin{center}

        \begin{small}
            \begin{sc}
                \begin{tabular}{lccccc}
                    \toprule
                    \multicolumn{1}{c}{Parameter}     &
                    \multicolumn{1}{c}{ MNIST}        &
                    \multicolumn{1}{c}{ FashionMNIST} &
                    \multicolumn{1}{c}{ ModelNet40}   &
                    \multicolumn{1}{c}{ ShapeNet10}   &
                    \multicolumn{1}{c}{ ScanNet10}                                                                                              \\
                    \midrule
                    HNet                              & [256, 256]      & [256, 256]      & [256]           & [256]           & [256]           \\
                    HNet-Heads                        & -               & -               & [256]           & [256]           & -               \\
                    MNet                              & [64, 64, 64]    & [64, 64, 64]    & [64, 128, 64]   & [128, 128, 128] & [32,32,32]      \\
                    dim($\bm{z}$)                     & 20              & 20              & 32              & 32              & 32              \\
                    HNet LR                           & $1e^{-4}$       & $1e^{-4}$       & $1e^{-4}$       & $1e^{-4}$       & $1e^{-4}$       \\
                    $\bm{z}$ LR                       & $1e^{-3}$       & $1e^{-3}$       & $1e^{-2}$       & $1e^{-2}$       & $1e^{-2}$       \\
                    batch size                        & 2048            & 2048            & 256             & 256             & 256             \\
                    Classifier                        & [128, 128, 128] & [128, 128, 128] & [128, 128, 128] & [128, 128, 128] & [128, 128, 128] \\
                    \bottomrule
                \end{tabular}
            \end{sc}
        \end{small}

    \end{center}
\end{table}

\section{Detailed Ablation Studies} \label{app:ablation}
\paragraph{Experimental setup and baselines}
To strictly control for variables, we employed a unified fixed baseline architecture (detailed in Table \ref{tab:baseline_config}) across all datasets and independent initialization trials.
This standardization ensures that observed deviations in reconstruction fidelity and classification accuracy are attributable solely to intrinsic data complexity or stochastic initialization variance, rather than network architectures or sizes.

\begin{table}[t]
    \caption{Baseline configuration for ablation and initialization experiments}
    \label{tab:baseline_config}
    \begin{center}

        \begin{small}
            \begin{sc}
                \begin{tabular}{lc}
                    \toprule
                    \multicolumn{1}{c}{ Hyperparameter} & \multicolumn{1}{c}{ Value} \\
                    \midrule
                    Latent Dim $z$                      & 20                         \\
                    MainNet Width                       & 64                         \\
                    MainNet Depth                       & 3                          \\
                    HyperNet Size                       & [256, 256]                 \\
                    HyperNet Heads                      & 0                          \\
                    Number of epochs                    & 500                        \\
                    Batch size                          & 1024                       \\
                    Classifier size                     & [128, 128, 128]            \\
                    Classifier batch size               & 128                        \\
                    Classifier epochs                   & 150                        \\
                    \bottomrule
                \end{tabular}
            \end{sc}
        \end{small}

    \end{center}
\end{table}

\paragraph{Initialization Stability}
Across the full spectrum of experimental configurations, we observed that the stochastic variance arising from random initialization is negligible relative to the magnitude of the primary performance metrics.
As shown in see Table \ref{tab:init_seeds}, quantitative analysis of five independent trials reveals consistently low standard deviations (typically on the order of $10^{-3}$ to $10^{-4}$), confirming that performance fluctuations remain tightly bounded around the mean.
This minimal variance certifies the reproducibility of our results and indicates that the proposed method possesses high robustness to initialization conditions.
Consequently, the observed performance deltas in our ablation studies can be confidently attributed to specific architectural interventions rather than stochastic noise, ensuring the statistical validity of our comparative analysis

\paragraph{Hyperparameter Sensitivity}
To rigorously evaluate the sensitivity of HyperINR, we implemented a comprehensive ablation protocol (detailed in Table~\ref{tab:baseline_config}) that mirrors the hyperparameter ranges established in prior baselines.
% , covering the latent sweeps used in \textit{Functa} and the architectural constraints of \textit{End-to-End INRs} and \textit{inr2vec}. 
Our empirical results demonstrate that HyperINR exhibits high stability across varying architectural configurations.
As shown in Table~\ref{tab:mnist_ablation} and \ref{tab:fmnist_ablation}, the model maintains high classification accuracy and low reconstruction error even when the latent dimension is significantly compressed (e.g., $dim(z)=20$) or when hypernetwork capacity is reduced.
This indicates that the structure encoded in weight space is intrinsic to the data structure rather than an artifact of over-parameterization.
Table~\ref{tab:fmnist_ablation} suggests that the optimal INR capacity depends on the complexity of the underlying data.
For MNIST, a compact INR architecture (width 64) is sufficient to achieve low reconstruction error ($2.40 \cdot 10^{-2}$) and high classification accuracy.
For FashionMNIST which is more complex than MNIST, we observe clear benefits from increased model capacity.
Scaling the main INR width to 128 results in improved reconstruction quality and higher downstream classification accuracy compared to smaller baselines. Similar patterns are observed when the depth of the main INR network is varied.
These findings indicate that HyperINR is robust and matches the network capacity to the data complexity.
Additionally, the hypernetwork needed to map the latent weight space to the weight space is relatively small when compared with other architectures~\ref{app:unifying_protocol}, and seems not to have a significant influence on reconstruction quality and classification accuracy. 
The above results highlight the robustness of our method and its strong performance even under minimal setups.

% ---------------------------------------------------------
% TABLE 2: INITIALIZATION STABILITY
% ---------------------------------------------------------
\begin{table}[t]
    \caption{Evaluations of Phase A across five random seeds}
    \label{tab:init_seeds}
    \begin{center}

        \begin{small}
            \begin{sc}
                \begin{tabular}{llll}
                    \toprule
                    \multicolumn{1}{c}{ Dataset}              &
                    \multicolumn{1}{c}{ Recon. Error (Train)} &
                    \multicolumn{1}{c}{ Recon. Error (Test)}  &
                    \multicolumn{1}{c}{ Class. Accuracy}                                                                                             \\
                    \midrule
                    MNIST                                     & $(2.42 \pm 0.08) \cdot 10^{-2}$ & $(3.32 \pm 0.11) \cdot 10^{-2}$ & $97.89 \pm 0.29$ \\
                    FashionMNIST                              & $(5.12 \pm 0.43) \cdot 10^{-2}$ & $(6.15 \pm 0.40) \cdot 10^{-2}$ & $86.82 \pm 0.98$ \\
                    \bottomrule
                \end{tabular}
            \end{sc}
        \end{small}

    \end{center}
\end{table}

\begin{table}[t]
    \caption{Evaluations of Phase B classifiers across five random seeds}
    \label{tab:init_seeds_cls}
    \begin{center}

        \begin{small}
            \begin{sc}
                \begin{tabular}{lll}
                    \toprule
                    \multicolumn{1}{c}{ Dataset}                 &
                    \multicolumn{1}{c}{ Classification Accuracy} &
                    \multicolumn{1}{c}{}                                              \\
                    \midrule
                    MNIST                                        & $98.12 \pm 0.06$ & \\
                    FashionMNIST                                 & $87.53 \pm 0.28$ & \\
                    \bottomrule
                \end{tabular}
            \end{sc}
        \end{small}

    \end{center}
\end{table}

% ---------------------------------------------------------
% TABLE: MNIST ABLATIONS
% ---------------------------------------------------------
\begin{table}[t]
    \caption{Ablation Study on MNIST.}
    \label{tab:mnist_ablation}
    \begin{center}

        \begin{small}
            \begin{sc}
                \begin{tabular}{lcccc}
                    \toprule
                    \multicolumn{1}{c}{ Hyperparameter}             &
                    \multicolumn{1}{c}{ Value}                      &
                    \multicolumn{1}{c}{ Reconstruction error Train} &
                    \multicolumn{1}{c}{ Reconstruction error Test}  &
                    \multicolumn{1}{c}{ Class. Accuracy}                                                                                 \\
                    \midrule
                    \multirow{7}{*}{Latent Dim $z$}                 & 10         & $5.88 \cdot 10^{-2}$ & $1.51 \cdot 10^{-1}$ & $81.01$ \\
                                                                    & 20         & $2.40 \cdot 10^{-2}$ & $3.24 \cdot 10^{-2}$ & $98.13$ \\
                                                                    & 64         & $8.63 \cdot 10^{-3}$ & $1.14 \cdot 10^{-2}$ & $97.44$ \\
                                                                    & 128        & $5.76 \cdot 10^{-3}$ & $6.86 \cdot 10^{-3}$ & $96.24$ \\
                                                                    & 256        & $5.42 \cdot 10^{-3}$ & $7.75 \cdot 10^{-3}$ & $91.40$ \\
                                                                    & 512        & $4.45 \cdot 10^{-3}$ & $6.35 \cdot 10^{-3}$ & $83.82$ \\
                                                                    & 1024       & $3.98 \cdot 10^{-3}$ & $7.17 \cdot 10^{-3}$ & $87.39$ \\
                    \midrule
                    \multirow{4}{*}{MainNet Width}                  & 32         & $2.81 \cdot 10^{-1}$ & $1.05 \cdot 10^{-1}$ & $97.24$ \\
                                                                    & 64         & $2.40 \cdot 10^{-2}$ & $3.24 \cdot 10^{-2}$ & $98.13$ \\
                                                                    & 128        & $1.88 \cdot 10^{-2}$ & $3.77 \cdot 10^{-2}$ & $97.42$ \\
                                                                    & 256        & $1.27 \cdot 10^{-2}$ & $5.49 \cdot 10^{-2}$ & $94.94$ \\
                    \midrule
                    \multirow{6}{*}{MainNet Depth}                  & 1          & $1.38 \cdot 10^{-1}$ & $1.49 \cdot 10^{-1}$ & $96.31$ \\
                                                                    & 2          & $3.24 \cdot 10^{-2}$ & $3.79 \cdot 10^{-2}$ & $98.22$ \\
                                                                    & 3          & $2.40 \cdot 10^{-2}$ & $3.24 \cdot 10^{-2}$ & $98.13$ \\
                                                                    & 4          & $3.61 \cdot 10^{-2}$ & $4.86 \cdot 10^{-2}$ & $96.50$ \\
                                                                    & 5          & $2.76 \cdot 10^{-2}$ & $5.35 \cdot 10^{-2}$ & $94.70$ \\
                                                                    & 6          & $2.17 \cdot 10^{-2}$ & $5.01 \cdot 10^{-2}$ & $97.36$ \\
                    \midrule
                    \multirow{3}{*}{HyperNet Size}                  & [128, 128] & $2.65 \cdot 10^{-2}$ & $3.97 \cdot 10^{-2}$ & $97.18$ \\
                                                                    & [256, 256] & $2.40 \cdot 10^{-2}$ & $3.24 \cdot 10^{-2}$ & $98.13$ \\
                                                                    & [512, 512] & $2.10 \cdot 10^{-2}$ & $3.36 \cdot 10^{-2}$ & $97.46$ \\
                    \midrule
                    \multirow{5}{*}{HyperNet Heads}                 & 0          & $2.40 \cdot 10^{-2}$ & $3.24 \cdot 10^{-2}$ & $98.13$ \\
                                                                    & 128        & $4.64 \cdot 10^{-2}$ & $5.29 \cdot 10^{-2}$ & $97.41$ \\
                                                                    & 256        & $2.63 \cdot 10^{-2}$ & $3.96 \cdot 10^{-2}$ & $94.41$ \\
                                                                    & 512        & $2.19 \cdot 10^{-2}$ & $3.73 \cdot 10^{-2}$ & $95.65$ \\
                                                                    & 1024       & $1.89 \cdot 10^{-2}$ & $3.22 \cdot 10^{-2}$ & $96.68$ \\
                    \bottomrule
                \end{tabular}
            \end{sc}
        \end{small}

    \end{center}
\end{table}

% ---------------------------------------------------------
% TABLE: FASHION MNIST ABLATIONS
% ---------------------------------------------------------
\begin{table}[t]
    \caption{Ablation Study on FashionMNIST.}
    \label{tab:fmnist_ablation}
    \begin{center}

        \begin{small}
            \begin{sc}
                \begin{tabular}{lcccc}
                    \toprule
                    \multicolumn{1}{c}{ Hyperparameter}            &
                    \multicolumn{1}{c}{ Value}                     &
                    \multicolumn{1}{c}{Reconstruction error Train} &
                    \multicolumn{1}{c}{ Reconstruction error Test} &
                    \multicolumn{1}{c}{ Class. Accuracy}                                                                                \\
                    \midrule
                    \multirow{7}{*}{Latent Dim $z$}                & 10         & $6.92 \cdot 10^{-2}$ & $1.09 \cdot 10^{-1}$ & $73.15$ \\
                                                                   & 20         & $5.72 \cdot 10^{-2}$ & $6.58 \cdot 10^{-2}$ & $86.23$ \\
                                                                   & 64         & $2.56 \cdot 10^{-2}$ & $3.17 \cdot 10^{-2}$ & $87.18$ \\
                                                                   & 128        & $1.91 \cdot 10^{-2}$ & $2.57 \cdot 10^{-2}$ & $83.52$ \\
                                                                   & 256        & $1.85 \cdot 10^{-2}$ & $2.70 \cdot 10^{-2}$ & $79.84$ \\
                                                                   & 512        & $1.52 \cdot 10^{-2}$ & $2.65 \cdot 10^{-2}$ & $64.01$ \\
                                                                   & 1024       & $1.41 \cdot 10^{-2}$ & $2.38 \cdot 10^{-2}$ & $72.82$ \\
                    \midrule
                    \multirow{4}{*}{MainNet Width}                 & 32         & $8.33 \cdot 10^{-2}$ & $9.50 \cdot 10^{-2}$ & $84.64$ \\
                                                                   & 64         & $5.72 \cdot 10^{-2}$ & $6.58 \cdot 10^{-2}$ & $86.23$ \\
                                                                   & 128        & $3.80 \cdot 10^{-2}$ & $5.50 \cdot 10^{-2}$ & $88.61$ \\
                                                                   & 256        & $2.29 \cdot 10^{-2}$ & $7.30 \cdot 10^{-2}$ & $83.64$ \\
                    \midrule
                    \multirow{6}{*}{MainNet Depth}                 & 1          & $1.13 \cdot 10^{-1}$ & $1.38 \cdot 10^{-1}$ & $75.29$ \\
                                                                   & 2          & $5.80 \cdot 10^{-2}$ & $7.38 \cdot 10^{-2}$ & $82.88$ \\
                                                                   & 3          & $5.72 \cdot 10^{-2}$ & $6.58 \cdot 10^{-2}$ & $86.23$ \\
                                                                   & 4          & $5.43 \cdot 10^{-2}$ & $6.79 \cdot 10^{-2}$ & $84.31$ \\
                                                                   & 5          & $4.60 \cdot 10^{-2}$ & $6.11 \cdot 10^{-2}$ & $88.11$ \\
                                                                   & 6          & $4.14 \cdot 10^{-2}$ & $6.29 \cdot 10^{-2}$ & $86.68$ \\
                    \midrule
                    \multirow{3}{*}{HyperNet Size}                 & [128, 128] & $5.13 \cdot 10^{-2}$ & $6.06 \cdot 10^{-2}$ & $86.60$ \\
                                                                   & [256, 256] & $5.72 \cdot 10^{-2}$ & $6.58 \cdot 10^{-2}$ & $86.23$ \\
                                                                   & [512, 512] & $3.97 \cdot 10^{-2}$ & $6.05 \cdot 10^{-2}$ & $83.65$ \\
                    \midrule
                    \multirow{5}{*}{HyperNet Heads}                & 0          & $5.72 \cdot 10^{-2}$ & $6.58 \cdot 10^{-2}$ & $86.23$ \\
                                                                   & 128        & $7.14 \cdot 10^{-2}$ & $1.17 \cdot 10^{-1}$ & $68.72$ \\
                                                                   & 256        & $5.00 \cdot 10^{-2}$ & $5.99 \cdot 10^{-2}$ & $88.07$ \\
                                                                   & 512        & $4.47 \cdot 10^{-2}$ & $8.18 \cdot 10^{-2}$ & $80.58$ \\
                                                                   & 1024       & $3.74 \cdot 10^{-2}$ & $7.88 \cdot 10^{-2}$ & $75.86$ \\
                    \bottomrule
                \end{tabular}
            \end{sc}
        \end{small}

    \end{center}
\end{table}

% \section{Additional experimental details on the comparison with Inr2Array}
% \label{app:inr2array}
% We compare our method against the baseline Inr2Array~\citep{zhou2023neural}, which originally reports an accuracy of 98.5\% on MNIST.
% Upon reproducing their method using the official code, we obtain a comparable 98.28\% accuracy.
% However, achieving this performance requires a Transformer-based classifier with approximately 22.3M parameters and a training time of $\approx$26 hours on a single NVIDIA A5000 (24GB RAM).
% HyperINR utilizes a significantly more compact architecture ($\approx$1M parameters), with the majority of weights concentrated in the hypernetwork heads required to project low-dimensional latent embeddings to the INR weight space. Consequently, our joint learning framework converges in under 30 minutes on the same hardware—representing a speedup of over $50\times$.

% We also evaluated Inr2Array without auxiliary strategies such as data augmentation, learning rate warmup, and weight decay. Under this controlled setting, Inr2Array's performance drops to 74.30\%. We did not scale down the Inr2Array model size, as Transformers typically benefit from over-parameterization; thus, this comparison favors the baseline.

\section{Matched Comparison Protocol}
\label{app:unifying_protocol}

\begin{table}[t]
    \caption{Architectural Comparison of INR-based Methods}
    \label{tab:inr_arch_comparison_merged}
    \begin{center}
        \begin{small}
            \begin{sc}
                \begin{tabular}{lcccc}
                    \toprule
                    \multicolumn{1}{c}{ Component} &
                    \multicolumn{1}{c}{ Ours}      &
                    \multicolumn{1}{c}{ INR2Array} &
                    \multicolumn{1}{c}{ NFN}       &
                    \multicolumn{1}{c}{ DWSNet}                                                                                                     \\
                    \midrule

                    SIREN INR                      & [64, 64, 64]    & [32, 32, 32]                       & [32, 32, 32]       & [32, 32]           \\

                    Decoder                        & [256, 256]      & 6 Transformer blocks               & [512, 512, 512]    & [32, 32, 32, 32]   \\

                    Latent $z$ dim                 & 20              & 512                                & 512                & 224                \\

                    Classifier                     & [128, 128, 128] & 12 Transformer blocks + [512, 512] & [1000, 1000, 1000] & Linear [224 to 10] \\
                    \bottomrule
                \end{tabular}
            \end{sc}
        \end{small}
    \end{center}
\end{table}

\begin{table}[ht]
    \caption{Matched comparison protocol 1: Ours vs Inr2Array and NFN (with Classification Accuracy)}
    \label{tab:inr_arch_comparison_no_dws}
    \begin{center}
        \begin{small}
            \begin{sc}
                \begin{tabular}{lccc}
                    \toprule
                    \multicolumn{1}{c}{ Module}    &
                    \multicolumn{1}{c}{ Ours}      &
                    \multicolumn{1}{c}{ INR2Array} &
                    \multicolumn{1}{c}{ NFN}                                                                                            \\
                    \midrule

                    SIREN INR                      & [32, 32, 32]    & [32, 32, 32]                       & [32, 32, 32]                \\

                    Decoder                        & [256, 256]      & 6 Transformer blocks               & [512, 512, 512] Equivariant \\

                    Latent $z$ Dim                 & 20              & 512                                & 512                         \\

                    Classifier                     & [128, 128, 128] & 12 Transformer blocks + [512, 512] & [1000, 1000, 1000]          \\
                    \midrule

                    MNIST (Acc)                    & 97.24           & 76.93                              & 61.27                       \\

                    FashionMNIST (Acc)             & 84.64           & 67.77                              & 57.29                       \\
                    \bottomrule
                \end{tabular}
            \end{sc}
        \end{small}
    \end{center}
\end{table}

\begin{table}[ht]
    \caption{Matched comparison protocol 2: Ours vs DWSNet (with Classification Accuracy)}
    \label{tab:inr_arch_comparison_HyperINR_dws_acc}
    \begin{center}
        \begin{small}
            \begin{sc}
                \begin{tabular}{lcc}
                    \toprule
                    \multicolumn{1}{c}{ Module} &
                    \multicolumn{1}{c}{ Ours}   &
                    \multicolumn{1}{c}{ DWSNet}                                            \\
                    \midrule

                    SIREN INR                   & [32, 32]              & [32, 32]         \\

                    Decoder                     & [256, 256]            & [32, 32, 32, 32] \\

                    Latent $z$ Dim              & 20                    & 224              \\

                    Classifier                  & 20 $\to$ 128 $\to$ 10 & 224 $\to$ 10     \\
                    \midrule

                    MNIST (Acc)                 & 97.09                 & 85.71            \\

                    FashionMNIST (Acc)          & 76.51                 & 67.06            \\
                    \bottomrule
                \end{tabular}
            \end{sc}
        \end{small}
    \end{center}
\end{table}

Table~\ref{tab:inr_arch_comparison_merged} details the components of all baseline models to highlight architectural differences and to justify the strategies used to ensure a fair comparison.

In Table~\ref{tab:inr_arch_comparison_no_dws}, we compare our HyperINR with two baseline INR-based methods, INR2Array and NFN. To ensure a fair comparison, we aligned our SIREN INR network with the baselines, while keeping the latent dimension and decoder configurations unchanged. This ensures that the core feature representations of each method remain unaffected—for instance, transformer-based decoders benefit from higher-dimensional latent embeddings. Additionally, we removed the data augmentation strategy used by INR2Array and NFN, where 10 independent INRs with different initializations were employed to enlarge the dataset. By doing so, we maintain a fair assessment of the intrinsic modeling capabilities of each architecture. Importantly, our work does not focus on architectural improvements; these design choices are made solely to demonstrate that the state-of-the-art results of HyperINR arise from our theoretical contributions rather than differences in network design. The classifiers employed by NFN and INR2Array have substantially more parameters than ours and were kept at their original scale in order to fairly assess the impact of excluding their dataset enlargement strategy.

When evaluating NFN and Inr2Array without their dataset enlargement strategy in a controlled setting, their classification accuracy drops significantly (for Inr2Aray MNIST: $76.93\%$, FashionMNIST: $67.77\%$, and for NFN MNIST: $61.27\%$, FashionMNIST: $57.29\%$), as depicted in Table~\ref{tab:inr_arch_comparison_no_dws}.

% The HyperINR achieves lower reconstruction error on MNIST (MSE $0.0227$) compared to Inr2Array ($0.0270$), and when the aforementioned data enlargement strategy is removed in the controlled setting, the reconstruction error of Inr2Array increases (MNIST train: 0.0769, MNIST test: 0.1414, FashionMNIST train: 0.0908, FashionMNIST test: 0.1171).

In Table~\ref{tab:inr_arch_comparison_HyperINR_dws_acc}, we present a direct comparison between HyperINR and DWSNet. For a fair evaluation, we aligned HyperINR to DWSNet by reducing the SIREN network size and classifier dimensions, resulting in almost the same total parameter count for both models. The decoder and latent dimensions were left unchanged to preserve the key innovations of each approach. This controlled setup ensures that the observed performance differences reflect the theoretical contributions of HyperINR rather than disparities in model capacity, highlighting the fairness and rigor of our comparative analysis. The hypernetwork we employ can be interpreted as a decoder when comparing our method to other models, since it maps latent weight representations $z$ to INR weights.

The above controlled comparisons highlight the robustness and efficiency of our method and demonstrate our theoretical model’s ability to achieve superior performance under a fair evaluation using a compact pipeline under minimal training enhancements.
%%%%%%%%%%%%%%%%%%%%%%%%%%%%%%%%%%%%%%%%%%%%%%%%%%%%%%%%%%%%%%%%%%%%%%%%%%%%%%%
%%%%%%%%%%%%%%%%%%%%%%%%%%%%%%%%%%%%%%%%%%%%%%%%%%%%%%%%%%%%%%%%%%%%%%%%%%%%%%%

\end{document}